\documentclass[lettersize,journal]{IEEEtran}

\usepackage[utf8]{inputenc}
\usepackage[T1]{fontenc}
\usepackage{amsmath,amsfonts,amssymb}
\usepackage{booktabs}
\usepackage{graphicx}
\usepackage{array}
\usepackage{textcomp}
\usepackage{stfloats}
\usepackage{url}
\usepackage[caption=false,font=footnotesize,labelfont=rm,textfont=rm]{subfig}
\usepackage{capt-of}
\usepackage{nicefrac}
\usepackage{microtype}
\usepackage{multirow}
\usepackage{diagbox}
\usepackage{siunitx}
\usepackage{float}
\usepackage{enumitem}
\usepackage[table]{xcolor}
\usepackage[normalem]{ulem}
\usepackage[ruled,vlined]{algorithm2e}
\usepackage{cite}
\usepackage{cuted}
\usepackage{makecell}

\usepackage[citecolor=blue, colorlinks]{hyperref}

\makeatletter
\def\input@path{{}{ieee/}{../}}
\makeatother
\graphicspath{{}{../}}

\definecolor{linecolor1}{RGB}{246, 248, 239}
\definecolor{linecolor2}{RGB}{230, 234, 217}
\definecolor{linecolor3}{RGB}{211, 222, 190}
\definecolor{oursrow}{HTML}{FFF8DE}

\newcolumntype{M}[1]{>{\centering\arraybackslash}m{#1}}

\newcommand{\mypara}[1]{\noindent\textbf{#1.}~}

\SetAlFnt{\small}
\SetAlCapFnt{\small}
\SetAlCapNameFnt{\small}
\SetAlCapHSkip{0pt}
\usepackage{soul}

\begin{document}

\title{CANIS: Generation-Assisted 3D Canonicalization via an Image-Semantic Bridge}

\author{Kendong Liu, Yuxin Yao, Junhui Hou
\thanks{This work was supported in part by the Natural Science Foundation of China under Grant 62422118, and in part by the Hong Kong Research Grants Council under Grants 11220426, 11219324, and N\_CityU1114/25.}
\thanks{The authors are with the Department of Computer Science, City University of Hong Kong, Hong Kong SAR, China (email: kdliu2-c@my.cityu.edu.hk; yuxinyao@cityu.edu.hk; jh.hou@cityu.edu.hk).}
}

\maketitle

\begin{strip}
\centering
\vspace{-2cm}
  \includegraphics[width=\textwidth]{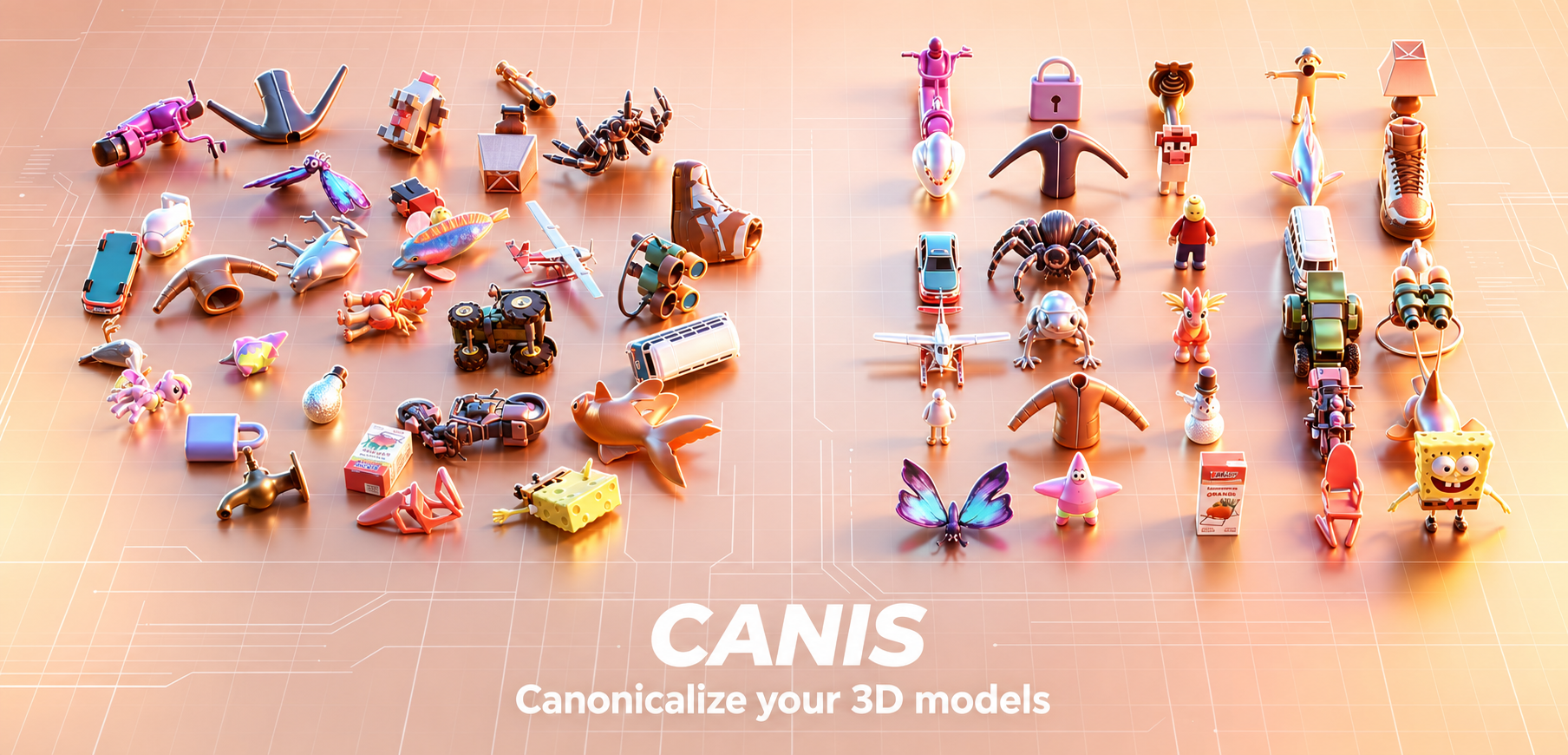}
  \vspace{-0.5cm}
  \captionof{figure}{Overview of CANIS on diverse 3D assets. Given objects in arbitrary orientations (left), CANIS estimates rotations that place them in a shared canonical frame (right). CANIS uses a frozen image to 3D generation prior and requires no canonicalization training or category templates specific to the evaluation benchmarks.}
  \label{fig:teaser}
\end{strip}

\begin{abstract}
Canonicalizing 3D object orientation is fundamental to 3D understanding and analysis. Existing approaches often rely on geometric cues, although 3D canonicalization ultimately requires a semantically meaningful orientation. To address this gap, we propose CANIS, a category-agnostic, generation-assisted framework that introduces the semantic orientation prior of a frozen image-to-3D generative model into 3D canonicalization, without canonicalization-specific training or category-specific templates. Specifically, CANIS first renders the input object from candidate viewpoints, selects an informative view, and generates a proxy in a canonical orientation. During generation, a sparse structural latent encoded from the input guides the proxy to preserve the geometry of an object. CANIS then uses the selected image as a semantic bridge between the input and the proxy. Image patches identify semantic regions on the proxy, and depth back-projection locates the corresponding regions on the input. The resulting semantic anchors constrain geometric matching, from which we estimate the rigid transformation that canonicalizes the input. 
Experiments on synthetic benchmarks validate CANIS and its key components, while qualitative results on partial observations and OmniObject3D suggest its applicability to incomplete and real-world scans. CANIS also improves downstream 3D classification, part segmentation, and dense correspondence under arbitrary rotations. Project page: \url{https://kenkenzaii.github.io/Canis}.
\end{abstract}


\section{Introduction}

\IEEEPARstart{3}{D} canonicalization estimates a rotation that maps an arbitrarily posed object to a consistent and semantically meaningful reference frame, typically \emph{upright} and \emph{front-facing}. A shared frame separates pose variation from intrinsic shape variation, allowing objects to be compared under consistent spatial coordinates. It therefore facilitates shape retrieval, correspondence estimation, deformation, annotation transfer, and cross-instance shape analysis. 

Canonicalization has been studied for category-level shape alignment~\cite{sun2021canonicalcapsules,sajnani2022_condor,agaram2023canonical}, upright estimation~\cite{pang2022upright,pang2025upright}, and semantic orientation disambiguation~\cite{scarvelissymmetry,wangorient}. Yet the task remains challenging because geometry alone may not determine a unique semantic orientation. Symmetric or repetitive structures can admit several geometrically plausible frames, while the cues that define upright and front-facing directions vary across object categories. These ambiguities are amplified in incomplete observations and real-world scans, where occlusion, missing regions, noise, and uneven sampling weaken the available geometric cues.

Existing methods designed specifically for 3D canonicalization primarily follow geometry-based or learning-based routes. Geometry-based methods infer canonical frames from symmetry, principal directions, bounding boxes, or registration cues~\cite{scarvelissymmetry}. Although directly grounded in 3D structure, these methods are sensitive to missing regions, noise, and approximate symmetries; moreover, geometrically plausible solutions may still confuse semantically distinct directions such as front and back. Learning-based methods encode equivariant representations, cross-instance correspondences, or category-level orientation priors from 3D data~\cite{spezialetti2020learning,sun2021canonicalcapsules,sajnani2022_condor,agaram2023canonical,pang2022upright,pang2025upright}. These methods improve orientation consistency, but often depend on canonicalization-specific training and category-structured data, which can limit generalization beyond the training distribution. 
Beyond these direct 3D approaches, image-based orientation estimators can be adapted to 3D canonicalization by applying them to rendered views and mapping their predictions back to object space~\cite{wangorient}. Although image appearance provides useful semantic cues, these methods operate primarily in the 2D image space and do not explicitly reason over the complete 3D structure of the input. Their predictions therefore remain sensitive to view selection and object visibility.
Recent 3D canonicalization approaches further incorporate vision-language reasoning or category-specific templates~\cite{jin2025one}. Although these strategies strengthen semantic disambiguation, they depend on the semantic coverage of external models or the availability of predefined category references. 
Collectively, existing approaches do not simultaneously provide semantic orientation reasoning, explicit grounding in the observed 3D structure, and category-agnostic operation without canonicalization-specific training or category-specific templates. Addressing this gap requires an instance-specific semantic orientation reference that can guide 3D canonicalization.

Recently, TRELLIS-OA~\cite{lu2025orientation}, an orientation-aligned image-to-3D generative model built upon the single-image-to-3D generative model TRELLIS~\cite{xiang2025structured}, learns semantic orientation regularities from consistently aligned 3D assets and generates image-conditioned shapes in a consistent canonical orientation. This kind of model offers a promising way to solve 3D canonicalization. 
Crucially, the value of this generative prior lies not only in producing a canonical proxy but also in the semantic link established through image-conditioned generation. Building on this insight, we propose CANIS, a generation-assisted framework that treats the generated shape as an instance-specific canonical proxy and uses the rendered conditioning image as a semantic bridge between an arbitrarily posed input and the proxy.
CANIS uses this connection to semantically constrain 3D correspondence matching, rather than predicting the orientation directly from the image. It then estimates the rotation that aligns the input with the proxy's canonical orientation and applies the rotation to the original input geometry. This formulation preserves the input shape and requires neither additional canonicalization-specific training nor category-specific templates.

Technically, CANIS realizes this framework through two complementary stages. Using the generated proxy for canonicalization requires addressing geometric discrepancies between the proxy and input, as well as geometrically plausible yet semantically incorrect correspondences during alignment. First, we select an informative view and incorporate input-derived structural guidance into proxy generation to obtain a geometrically compatible canonical proxy. The corresponding RGB-D observation then establishes a semantic bridge between the input and proxy, from which paired semantic anchors are constructed to constrain 3D correspondence matching and recover the canonicalizing rotation.

We evaluate CANIS on synthetic benchmarks, partial observations, and real-world scans, with ablations isolating the contributions of shape guidance and semantic anchors. We further demonstrate its benefits for downstream 3D classification, part segmentation, and dense correspondence.

In summary, the main contributions of this paper are:

\begin{itemize}[itemsep=1pt, topsep=0pt, leftmargin=15pt]
    \item We introduce CANIS, a novel generation-assisted framework that reformulates semantic 3D canonicalization as alignment with a generator-produced canonical proxy, turning orientation-aligned 3D generation into an actionable semantic prior without canonicalization-specific training or category-specific templates.
    
    \item We bridge generation and registration through two complementary stages. The first uses input shape guidance to synthesize a geometrically compatible canonical proxy. The second uses semantic anchors to constrain 3D correspondence matching and recover the canonicalizing rotation despite geometric discrepancies between the proxy and input.
    
    \item Extensive experiments on complete synthetic objects, partial observations, and real-world scans validate CANIS and its key components, while evaluations on 3D classification, part segmentation, and dense correspondence demonstrate the downstream value of generation-assisted canonicalization.
    
\end{itemize}

\vspace{-0.2cm}
\section{Related Work}
\label{sec:related}

\subsection{3D Canonicalization}

Prior work addresses canonical orientation through category-level alignment~\cite{gu2020weakly,sun2021canonicalcapsules, sajnani2022_condor,agaram2023canonical,di2024shapematcher, jin2026canoverse}, upright estimation~\cite{pang2022upright,pang2025upright}, and full-orientation or front-facing disambiguation~\cite{spezialetti2020learning,scarvelissymmetry, wangorient}. Geometry-based approaches rely on symmetry axes, principal directions, oriented bounding boxes, or template registration. Such cues become unreliable under partial observations, noise, and geometric symmetries.

Learning-based category-level methods replace handcrafted cues with learned shape priors or equivariant representations. CaCa~\cite{sun2021canonicalcapsules} learns semantic capsules and keypoints from randomly rotated point clouds, while ConDors~\cite{sajnani2022_condor} uses rotation-equivariant features to canonicalize full and partial shapes. Canonical Fields~\cite{agaram2023canonical} operates on pretrained neural fields, and ShapeMatcher~\cite{di2024shapematcher} jointly learns canonicalization, segmentation, retrieval, and deformation using a precanonicalized source database. Gu et al.~\cite{gu2020weakly} jointly estimated a canonical shape and its pose from multiple partial observations. These methods require task-specific training on category-structured data: CaCa and ConDor are trained on individual or fixed sets of ShapeNet categories, while Canonical Fields and ShapeMatcher assume pretrained NeRFs or reference shape collections. In contrast, we propose a category-agnostic framework that leverages a frozen orientation-aligned generator as a semantic prior and \textit{does not} require canonicalization-specific training.

Methods~\cite{pang2025upright,pang2022upright,scarvelissymmetry} that estimate semantic orientation axes differ in the components of the canonical frame they recover and the supervision used to define those axes. UprightNet~\cite{pang2022upright} and UprightNet+~\cite{pang2025upright} estimate an object's supporting base and recover its upright direction, whereas SR3D~\cite{scarvelissymmetry} predicts the complete orientation axes with a symmetry-aware formulation. Jin et al.~\cite{jin2025one} combined geometric consistency with vision-language cues but require one canonical reference template per category. CANIS instead constructs instance-specific semantic anchors between the input and a generated canonical proxy, without a category-specific reference model or a learned axis regressor.

\subsection{Image-based Object Orientation Estimation}

Image-based orientation models~\cite{wangorient, wang2026orient} infer semantic object axes from RGB observations rather than directly canonicalizing 3D geometry. OrientAnything~\cite{wangorient} predicts distributions over three orientation angles from single images using a model trained on rendered 3D assets. OrientAnythingV2~\cite{wang2026orient} extends this formulation with symmetry-aware orientation distributions and multi-view relative rotation estimation. Both methods infer orientation from image features without enforcing consistency with the input 3D geometry. Their accuracy depends on the rendered view and the visibility of parts that define the object's front or upright direction; changes in texture, lighting, and rendering style can weaken these cues. For evaluation, we transform their camera relative predictions into the object frame using the known camera pose. 

CANIS instead synthesizes a canonical proxy from the selected rendering and uses the corresponding RGB-D image as a semantic bridge to establish paired anchors on the input and proxy. The subsequent registration stage uses these anchors to constrain geometric correspondence search and solves for the rotation that maps the input to the proxy's canonical frame, introducing an explicit 3D geometric constraint absent from direct image-based orientation prediction.

\begin{figure*}
    \centering
    \vspace{\baselineskip}
    \includegraphics[width=1\linewidth]{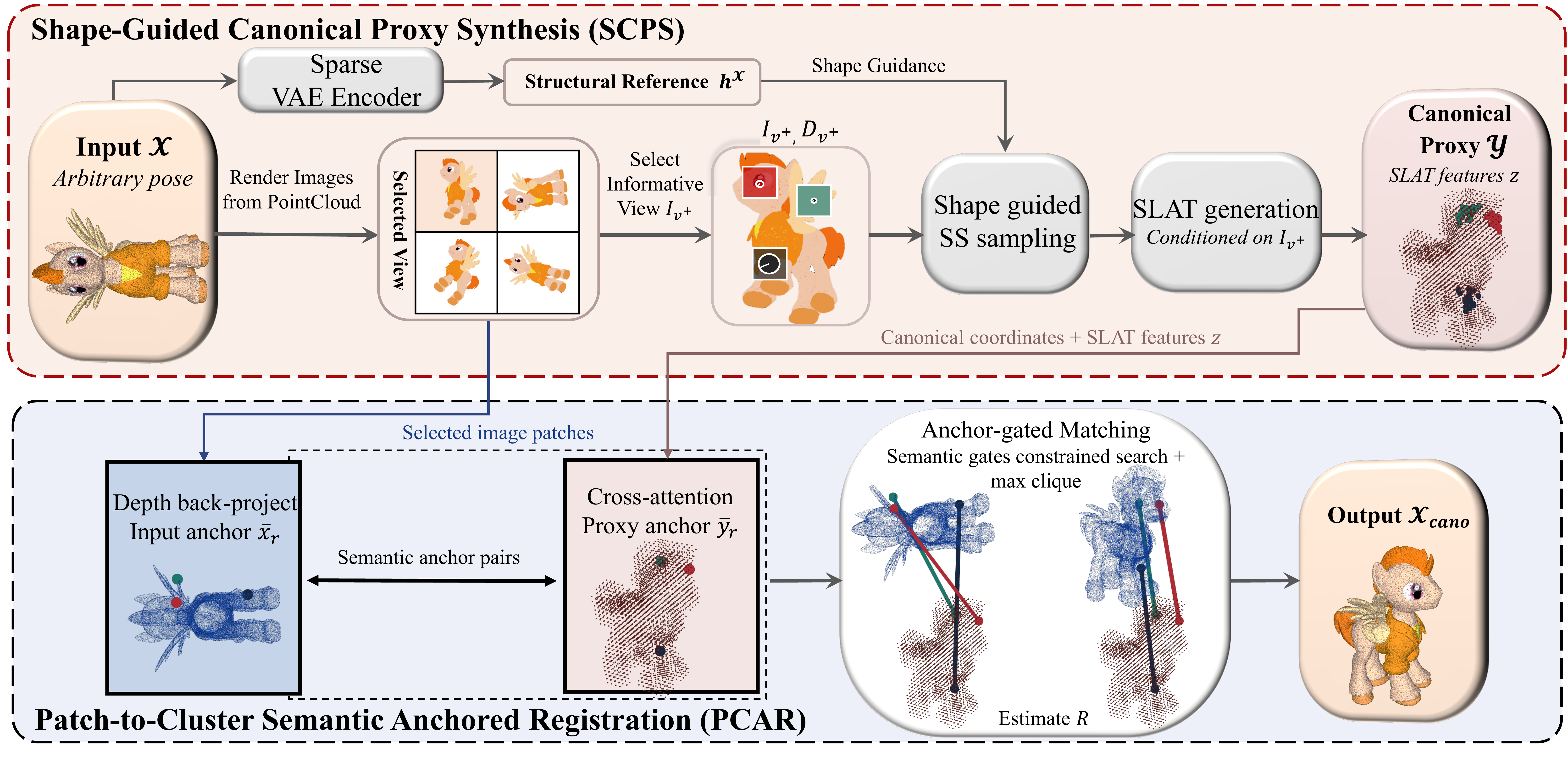}
    \vspace{-0.7cm}
    \caption{\textbf{Pipeline of CANIS.}  SCPS selects an informative rendering and synthesizes a canonical proxy using structural guidance from the input. PCAR uses the selected image and its depth map to construct paired semantic anchors on the input and proxy. These anchors restrict geometric correspondence search, from which CANIS estimates the rotation applied to the original input.}
    \label{fig:pipeline}
\end{figure*}

\subsection{3D Registration and Correspondence Estimation}

Rigid registration is commonly solved by alternating correspondence assignment and transformation estimation, as in ICP~\cite{besl1992method} and its variants~\cite{segal2009generalized,rusinkiewicz2019symmetric,zhang2021fast}. Feature-based pipelines instead match local descriptors and apply outlier rejection before estimating the rigid transformation. Representative descriptors include the handcrafted Fast Point Feature Histograms (FPFH)~\cite{rusu2009fast} and the learned Fully Convolutional Geometric Features (FCGF)~\cite{choy2019fully}. 
Recent approaches either regress transformations from global features~\cite{aoki2019pointnetlk}, learn soft correspondences or latent mixture assignments~\cite{wang2019deep,yuan2020deepgmr}, or use transformers and graph reasoning to improve correspondence estimation~\cite{qin2022geometric,zhang20233d,lepard2021}.  
Across these formulations, reliable registration still depends on sufficient geometric compatibility between the source and target. The generated proxy in CANIS may also differ from the input in local structure, proportions, or missing regions, further complicating correspondence estimation. The subsequent registration stage retains geometric descriptors but uses image-induced semantic anchors to restrict FPFH-based or FCGF-based matching to semantically aligned neighborhoods before pose estimation.

\subsection{3D Generative Priors}

Image-conditioned 3D generators reconstruct or synthesize 3D assets from one or more images using feed-forward reconstruction, diffusion, or structured latent representations~\cite{hong2024lrm,zhang2024clay,tao2025gsv3d,xiang2025structured,xiang2025trellis2,liu2025acc3d,lu2025orientation}. Although these models provide strong 3D priors, their outputs do not necessarily follow a consistent canonical orientation. Intermediate representations learned by such generators have also been used for semantic correspondence and shape interpolation~\cite{du2025hierarchical,sun2026morphany3d}.

TRELLIS-OA~\cite{lu2025orientation} introduces orientation-aligned 3D generation and fine-tunes generative backbones, including TRELLIS~\cite{xiang2025structured}, on the curated Objaverse-OA~\cite{lu2025orientation} dataset to produce consistently oriented assets. Its objective is to synthesize a new orientation-aligned object rather than canonicalize an existing 3D input. The generated asset may therefore differ from the input in unobserved geometry and local structure. CANIS uses TRELLIS-OA only as an orientation prior: the synthesis stage constrains the generated proxy toward the observed input structure, while the registration stage aligns the original geometry with the proxy's canonical orientation.

\section{Proposed Method}

Let \(\mathcal{X}=\{\mathbf{x}_n\}_{n=1}^{N_x} \subset\mathbb{R}^{3}\) denote a centered input point cloud in an arbitrary orientation. Our goal is to estimate a rotation \(\mathbf{R}\in\mathrm{SO}(3)\) that maps the input to a shared upright and front-facing frame:
\begin{equation}
    \mathcal{X}_{\mathrm{can}}
    =
    \left\{
        \mathbf{R}\mathbf{x}_n
    \right\}_{n=1}^{N_x}.
    \label{eq:canonicalization_target}
\end{equation}

CANIS introduces a new generation-assisted framework for semantic 3D canonicalization by exploiting the semantic orientation priors encoded in a pretrained orientation-aligned single-image-to-3D generator~\cite{lu2025orientation} to construct an instance-specific canonical reference. A generative orientation prior alone is insufficient for canonicalization; its output must be geometrically compatible with the observed 3D input and semantically grounded in the input geometry through the conditioning image. CANIS couples input-guided proxy synthesis with a shared conditioning image that serves as a semantic bridge between the observed and generated 3D structures. This design grounds the canonical prior in the original geometry and enables semantically constrained 3D correspondence matching.
As illustrated in Fig.~\ref{fig:pipeline}, CANIS comprises shape-guided canonical proxy synthesis (SCPS) and patch-to-cluster semantic anchor registration (PCAR). SCPS (Sec.~\ref{sec:scps}) produces a geometry-compatible canonical proxy \(\mathcal{Y}\); PCAR (Sec.~\ref{sec:PCAR}) then uses semantic anchors to constrain 3D correspondence matching and estimate the canonicalizing rotation \(\mathbf{R}\), which is applied to the original input geometry. All pretrained components remain frozen.

\subsection{Orientation-Aligned Image-to-3D Generation}
\label{sec:generator_background}

CANIS uses two properties of orientation-aligned image-to-3D generation: the generated geometry follows a consistent semantic orientation, and the generation process relates image tokens to 3D latent regions through cross-attention. We instantiate CANIS with TRELLIS-OA~\cite{lu2025orientation}, an orientation-aligned version of TRELLIS~\cite{xiang2025structured}. TRELLIS-OA first samples a sparse-structure (SS) latent and decodes it into active voxel centers. It then assigns image-conditioned structured latent (SLAT) features to these voxels. We denote the frozen SS encoder and decoder by \(\mathrm{E}\) and \(\mathrm{D}\), respectively, and the SLAT generator by \(\mathrm{G}\). During SLAT sampling, the proxy voxels attend to the conditioning-image tokens. CANIS records this native voxel-to-image-token cross-attention for subsequent semantic association, without training an additional cross-modal matcher.

\subsection{Shape-Guided Canonical Proxy Synthesis}
\label{sec:scps}

Shape-guided canonical proxy synthesis (SCPS) aims to produce a proxy that inherits the generator's semantically canonical orientation while retaining sufficient geometric compatibility with the input for subsequent registration. Because the generator observes only a single rendering, it must infer self-occluded geometry, and the resulting proxy may differ from the input in scale, local structure, or part configuration. SCPS therefore combines informative-view conditioning with structural guidance from the observed input. The selected image supplies the semantic orientation prior, whereas the structural reference limits instance-specific geometric drift.

\vspace{0.5em}
\mypara{Informative View Selection}
For an arbitrarily rotated input, candidate views can differ markedly in visible support and orientation-discriminative evidence; in some views, much of the object may be self-occluded or the visible regions may provide only weak semantic cues. Although these renderings are geometrically consistent observations of the same 3D shape, the adopted generator accepts only one conditioning image at a time. Switching the conditioning image across sampling steps would change the observed surface associated with each image-token index, making cross-step voxel-to-token attention semantically inconsistent. Alternatively, generating one proxy per view and aggregating the candidates would incur repeated inference, while their proxy supports and view-specific attention maps are not pointwise aligned. We therefore select one informative view and use it consistently for proxy synthesis and subsequent semantic grounding.

We render \(\mathcal{X}\) from \(V\) candidate viewpoints. Each view \(v\) provides an RGB image \(\mathbf{I}_v\in\mathbb{R}^{H\times W\times 3}\), a depth map \(\mathbf{D}_v\in\mathbb{R}^{H\times W}\), a foreground mask \(\mathbf{M}_v\in\{0,1\}^{H\times W}\), and camera parameters \(\mathbf{C}_v\), including the intrinsics and extrinsics. 
Let \(\boldsymbol{\phi}_v\in\mathbb{R}^{d_\phi}\) collect the normalized evidence cues computed from \(\mathbf{I}_v\) and \(\mathbf{M}_v\), and let \(\mathbf{w}\in\mathbb{R}^{d_\phi}\) denote the corresponding fixed cue-weight vector. We select
\begin{equation}
    v^\dagger
    =
    \underset{1\leq v\leq V}
    {\operatorname{arg\,max}}\;
    \mathbf{w}^{\top}\boldsymbol{\phi}_v,
    \label{eq:view_select}
\end{equation}
where the components of \(\boldsymbol{\phi}_v\) measure foreground coverage, contour information, appearance variation within the foreground region, and contrast between the foreground and background.

We denote the selected observation by
\[
    (\mathbf{I}^\dagger,\mathbf{D}^\dagger,
     \mathbf{M}^\dagger,\mathbf{C}^\dagger)
    =
    (\mathbf{I}_{v^\dagger},\mathbf{D}_{v^\dagger},
     \mathbf{M}_{v^\dagger},\mathbf{C}_{v^\dagger}).
\] The selected image \(\mathbf{I}^\dagger\) conditions all proxy-generation steps, ensuring that the recorded attention maps share a common image-token reference. Its paired depth map, foreground mask, and camera parameters are retained so that PCAR can localize the same image regions on the observed input surface.

\vspace{0.5em}
\mypara{Shape-Guided SS Sampling}
An informative conditioning view improves semantic orientation cues but cannot fully constrain geometry that is not visible in the selected rendering. We therefore encode the available input geometry in the SS latent space as a structural reference:
\begin{equation}
    \mathbf{h}^{\mathcal{X}}
    =
    \mathrm{E}
    \left(
        \operatorname{Vox}(\mathcal{X})
    \right),
    \qquad
    \mathbf{h}^{\mathcal{X}}
    \in
    \mathcal{H}_{\mathrm{SS}}.
    \label{eq:ss_reference}
\end{equation}

Let \(\Phi_s(\mathbf{h}_s\mid\mathbf{I}^{\dagger})\) denote the \(s\)-th image conditioned update of the frozen SS sampler. Starting from Gaussian noise \(\mathbf{h}_0\in\mathcal{H}_{\mathrm{SS}}\), each sampling step combines this update with the structural reference \(\mathbf{h}^{\mathcal X}\):
\begin{equation}
\begin{aligned}
    \mathbf{h}_{s+1}
    &=
    (1-\alpha_s)
    \Phi_s
    \left(
        \mathbf{h}_s
        \mid
        \mathbf{I}^{\dagger}
    \right)
    +
    \alpha_s\mathbf{h}^{\mathcal X},\\
    \alpha_s
    &=
    \alpha
    \left(
        1-\frac{s}{T}
    \right),
    \qquad
    s=0,\ldots,T-1,
\end{aligned}
\label{eq:ss_guidance}
\end{equation}
where \(T\) is the number of SS sampling updates and \(\alpha\in[0,1]\) is the initial guidance strength.

Although $\mathbf{h}_X$ retains the observed input orientation, it is injected only as a weak, decaying structural reference rather than a pose target. Since $\alpha_s \leq 0.10$, the image-conditioned update remains 
dominant throughout sampling. Subsequent conditional updates preserve reference structures consistent with $\mathbf{I}^{\dagger}$ while suppressing incompatible orientation-specific components, thereby improving proxy fidelity without overriding the generator's canonical-orientation prior.

The final latent is decoded into the active voxel support of the canonical proxy:
\begin{equation}
    \mathcal{Y}
    =
        \mathrm{D}(\mathbf{h}_T)
    =
    \{\mathbf{y}_i\}_{i=1}^{N_y},
    \qquad
    \mathbf{y}_i\in\mathbb{R}^{3}.
    \label{eq:proxy_support}
\end{equation}
The SLAT generator assigns an image-conditioned feature to each proxy voxel:
\begin{equation}
    \mathcal{Z}
    =
    \mathrm{G}
    \left(
        \mathcal{Y}
        \mid
        \mathbf{I}^{\dagger}
    \right)
    =
    \left\{
        (\mathbf{y}_i,\mathbf{z}_i)
    \right\}_{i=1}^{N_y},
    \qquad
    \mathbf{z}_i\in\mathbb{R}^{d_z}.
    \label{eq:proxy_slat}
\end{equation}
During SLAT sampling, we record the conditional voxel-to-image-token cross-attention. SCPS passes the canonical proxy, the selected RGB-D observation and camera, and the recorded attention maps to PCAR.

\subsection{Patch-to-Cluster Semantic Anchor Registration}
\label{sec:PCAR}

\begin{figure*}[t]
    \centering
    \vspace{\baselineskip}
    \includegraphics[width=0.98\linewidth]{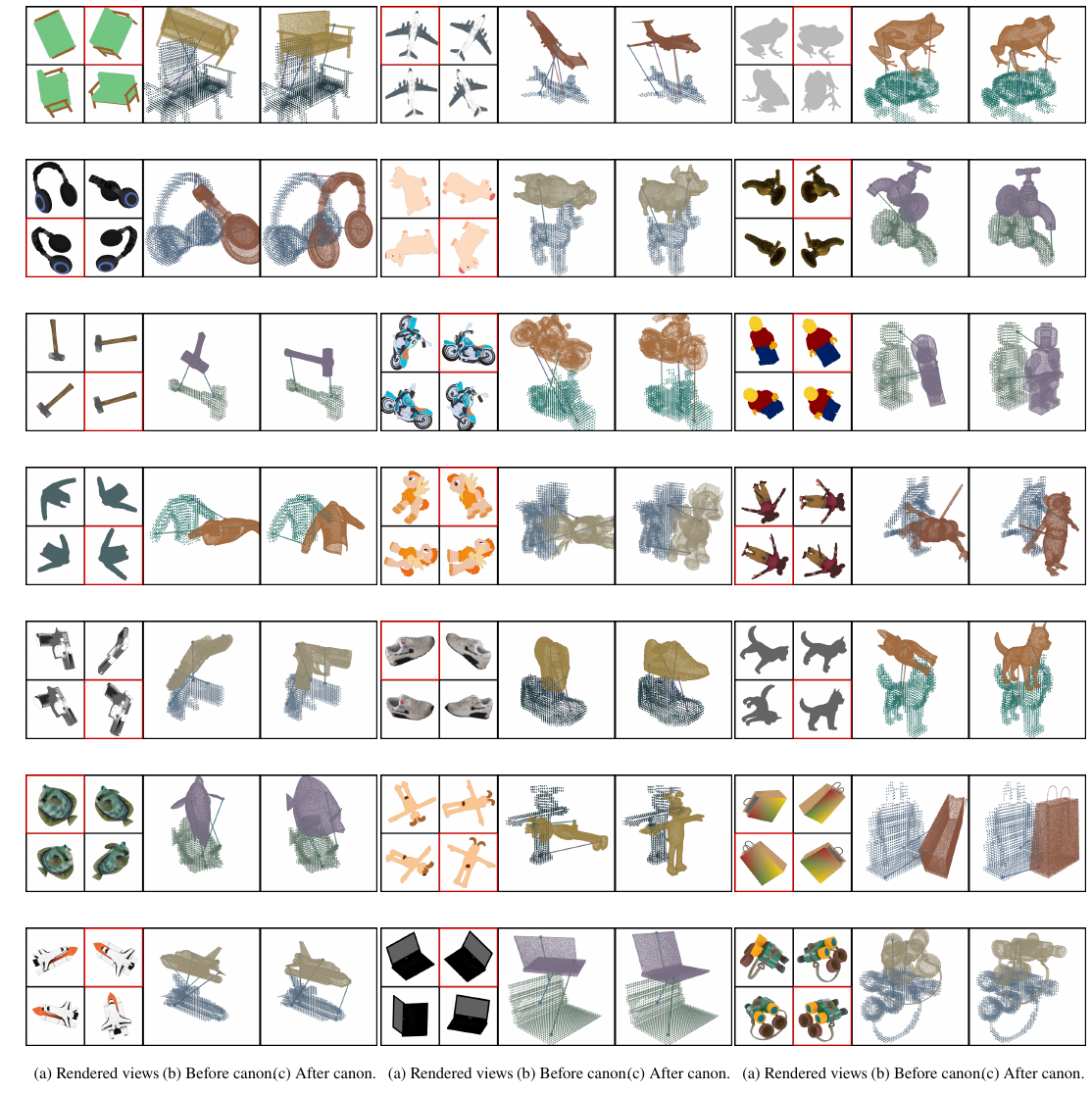}
    \vspace{-0.5cm}
    \caption{
    Visualization of informative view selection, key semantic anchor pairs, and canonicalization results. (a) Candidate rendered views, with the selected informative views outlined in \textcolor{red}{red box}; only key semantic anchor pairs are shown for clarity. (b) Input shape and proxy before canonicalization. (c) Canonicalized results.}
    \vspace{-0.4cm}
    \label{fig:3d3d_grid}
\end{figure*}

Even with a geometry-compatible proxy, large initial rotations, repeated or approximately symmetric parts, and residual proxy--input discrepancies can produce geometrically plausible but semantically incorrect correspondences. Patch-to-cluster semantic anchor registration (PCAR) resolves this ambiguity by using the selected RGB-D observation as a semantic bridge between the input and the canonical proxy: depth and camera parameters localize image regions on the observed input surface, while generation-time cross-attention associates the same regions with the proxy. PCAR first constructs paired semantic anchors through patch-to-cluster association and then uses these anchors to constrain local 3D descriptor matching and estimate the canonicalizing rotation. Fig.~\ref{fig:3d3d_grid} displays some semantic anchor pairs, together with the selected informative views and corresponding canonicalization results. 

\vspace{0.5em}
\noindent{\textit{\textbf{1) Patch-to-Cluster Association}}}
\leavevmode\par\nobreak

\vspace{0.25em}
Individual token--voxel attention responses may be spatially diffuse and sensitive to local proxy--input discrepancies. PCAR therefore aggregates them into regional units: an image block defines a 2D patch, and a spatial block of proxy voxels defines a 3D voxel cluster. Representative patch to cluster associations are visualized in Fig.~\ref{fig:2d_3d}.

\vspace{0.5em}
\mypara{Image Block Construction}
We apply the same resizing and cropping operations to the selected image and its foreground mask. The frozen image encoder partitions the processed image into a regular grid of \(N_p\) spatial tokens. Let \(\mathcal{P}_j\) denote the set of pixels assigned to token \(j\) according to this grid, excluding any nonspatial special tokens. We define the foreground occupancy of token \(j\) as
\begin{equation}
    \rho_j
    =
    \frac{1}{|\mathcal{P}_j|}
    \sum_{\mathbf{u}\in\mathcal{P}_j}
    \mathbf{M}^{\dagger}(\mathbf{u}),
    \label{eq:foreground_occupancy}
\end{equation}
where \(\mathbf{u}\) is a pixel coordinate. We retain tokens satisfying \(\rho_j\geq\tau_{\mathrm{fg}}\) and organize adjacent retained tokens into regular, non-overlapping image blocks \(\{\mathcal{B}_b\}_{b=1}^{N_b}\). Each \(\mathcal{B}_b\) is a set of token indices, and \(\mathbf{u}_b\) denotes its center in image coordinates. Regular blocking preserves the image-plane layout without requiring a separate keypoint detector.

\vspace{0.5em}
\mypara{Cross Attention Aggregation}
To obtain stable semantic associations between the canonical proxy and the selected image, we aggregate the conditional cross attention recorded across sampling steps, network layers, and attention heads. The resulting attention matrix relates each proxy voxel to the image tokens and supports the subsequent construction of semantic anchors.
For a single attention head \(h\), let
\(\mathbf{A}^{(s,\ell,h)}\in[0,1]^{N_y\times N_p}\)
denote the attention map recorded at sampling step \(s\) and network layer \(\ell\), where \(N_y\) and \(N_p\) are the numbers of proxy voxels and spatial image tokens, respectively. We aggregate the retained attention maps as
\vspace{-0.1cm}
\begin{equation}
    \mathbf{A}
    =
    \frac{1}{|\Omega|N_h}
    \sum_{(s,\ell)\in\Omega}
    \sum_{h=1}^{N_h}
    \mathbf{A}^{(s,\ell,h)},
    \qquad
    \mathbf{A}=[a_{ij}]
    \in[0,1]^{N_y\times N_p}.
    \label{eq:attention_aggregation}
\end{equation}
where \(\Omega\) is the set of retained sampling step and network layer pairs, and \(N_h\) is the number of attention heads. 

Each entry \(a_{ij}\) represents the mean attention response of proxy voxel \(\mathbf{y}_i\) to image token \(j\), aggregated over the retained sampling steps, network layers, and attention heads. In practice, \(\Omega\) comprises later sampling steps and deeper network layers.

Cross-attention relates image tokens to proxy voxels but is not visibility-aware. Before block association, we apply silhouette-consistent candidate filtering to identify valid proxy support. Specifically, we render $\mathcal{Y}$ from a discrete set of candidate views $\Theta$. For each $\theta\in\Theta$, $\Pi_{\theta}$ denotes the corresponding 3D-to-2D projection, and $\mathbf{M}^{\mathcal{Y}}_{\theta}$ is the resulting binary foreground mask. We select the projection with the largest silhouette overlap:
\vspace{-0.1cm}
\begin{equation}
    \theta^\star
    =
    \underset{\theta\in\Theta}{\operatorname{argmax}}
    \;
    \operatorname{IoU}
    \left(
        \mathbf{M}^{\mathcal{Y}}_{\theta},
        \mathbf{M}^{\dagger}
    \right).
    \label{eq:silhouette_projection_selection}
\end{equation}

Let $\mathcal{F}^{\dagger}=\{\mathbf{u}\mid\mathbf{M}^{\dagger}(\mathbf{u})=1\}$ denote the foreground region of the selected input image. We retain the proxy voxels whose projections fall within this region:
\begin{equation}
    \mathcal{I}_{\mathrm{fg}}
    =
    \left\{
        i\in\{1,\ldots,N_y\}
        \,\middle|\,
        \Pi_{\theta^\star}(\mathbf{y}_i)
        \in\mathcal{F}^{\dagger}
    \right\}.
    \label{eq:foreground_proxy_support}
\end{equation}
The set $\mathcal{I}_{\mathrm{fg}}$ defines the proxy candidates for block association without modifying the proxy coordinates or cross-attention matrix $\mathbf{A}$. The camera rotation associated with $\theta^\star$ is not used to initialize or compose the final transformation. The binary masks provide only visibility filtering; semantic associations and the final transformation are obtained from cross-attention and anchor-constrained 3D matching, respectively.

We partition the retained proxy indices into regular, non-overlapping spatial blocks \(\{\mathcal{G}_g\}_{g=1}^{N_g}\), where \(\mathcal{G}_g\subseteq\mathcal{I}_{\mathrm{fg}}\). The association between a proxy block \(\mathcal{G}_g\) and an image block \(\mathcal{B}_b\) is defined as the mean cross-attention over their voxel--token pairs:
\begin{equation}
    s_{gb}
    =
    \frac{1}{|\mathcal{G}_g|\,|\mathcal{B}_b|}
    \sum_{i\in\mathcal{G}_g}
    \sum_{j\in\mathcal{B}_b}
    a_{ij}.
    \label{eq:block_association}
\end{equation}
Averaging over the two blocks normalizes their occupancies and reduces sensitivity to isolated attention responses.

To obtain spatially distributed associations, we apply farthest-point sampling to the image-block centers \(\{\mathbf{u}_b\}_{b=1}^{N_b}\) and denote the selected indices, in sampling order, by \(\{b_r\}_{r=1}^{N_{\mathrm{anc}}}\), where \(N_{\mathrm{anc}}\leq\min(N_b,N_g)\). Each selected image block \(\mathcal{B}_{b_r}\) is then assigned to its highest-scoring previously unassigned proxy block \(\mathcal{G}_{g_r}\), yielding
\[
    \left\{
        (\mathcal{B}_{b_r},\mathcal{G}_{g_r})
    \right\}_{r=1}^{N_{\mathrm{anc}}}.
\]
The one-to-one assignment prevents multiple image blocks from collapsing onto the same proxy region.

\vspace{0.5em}
\noindent{\textit{\textbf{2) Semantic Anchor Construction and Registration}}}
\leavevmode\par\nobreak

\vspace{0.25em}
The patch-to-cluster pairs obtained above encode regional semantic affinity and require 3D localization before they can constrain geometric registration.

\vspace{0.5em}
\mypara{Semantic Anchor Construction}
For each selected block pair \((\mathcal{B}_{b_r},\mathcal{G}_{g_r})\), PCAR defines an input-side anchor and a proxy-side anchor.

On the input side, \(\widehat{\mathbf{u}}_r\) is set to the block center \(\mathbf{u}_{b_r}\) when valid foreground depth is available at that location, and otherwise to the nearest pixel with valid foreground depth. The corresponding input-side anchor is obtained by depth-conditioned inverse projection:
\begin{equation}
    \bar{\mathbf{x}}_r
    =
    \Pi_{\mathbf{C}^{\dagger}}^{-1}
    \left(
        \widehat{\mathbf{u}}_r,\,
        \mathbf{D}^{\dagger}
        (\widehat{\mathbf{u}}_r)
    \right).
    \label{eq:input_anchor}
\end{equation}
Here, \(\Pi_{\mathbf{C}^{\dagger}}^{-1}\) maps a pixel--depth pair to the input coordinate frame under the selected camera.

On the proxy side, the response of proxy voxel \(\mathbf{y}_i\) to the
corresponding image block is: 
\begin{equation}
    w_{ir}
    =
    \frac{1}{|\mathcal{B}_{b_r}|}
    \sum_{j\in\mathcal{B}_{b_r}}
    a_{ij},
    \qquad
    i\in\mathcal{G}_{g_r}.
    \label{eq:anchor_weight}
\end{equation}
The proxy-side anchor is then defined as the attention-weighted centroid of the associated proxy block:
\begin{equation}
    \bar{\mathbf{y}}_r
    =
    \frac{
        \sum_{i\in\mathcal{G}_{g_r}}
        w_{ir}\mathbf{y}_i
    }{
        \sum_{i\in\mathcal{G}_{g_r}}w_{ir}
    },
    \label{eq:proxy_anchor}
\end{equation}
The resulting semantic-anchor pairs are collected as
\begin{equation}
    \mathcal{A}_{\mathrm{sem}}
    =
    \left\{
        \left(
            \bar{\mathbf{x}}_r,
            \bar{\mathbf{y}}_r
        \right)
    \right\}_{r=1}^{N_a}.
    \label{eq:semantic_anchor_set}
\end{equation}
Because both anchors are induced by the same image block, each pair links semantically associated neighborhoods on the observed input and the canonical proxy.

\vspace{0.5em}
\mypara{Anchor-Constrained Registration}
Before descriptor extraction, we align the center and isotropic scale of the proxy with those of the input.
Let \(\mathbf{c}_x,\mathbf{c}_y\) and \(\delta_x,\delta_y>0\) denote their respective centers and geometric scales. We define a normalization map for any proxy coordinate \(\mathbf{p}\in\mathbb{R}^{3}\):
\begin{equation}
    \mathcal{N}_{y}(\mathbf{p})
    =
    \frac{\delta_x}{\delta_y}
    (\mathbf{p}-\mathbf{c}_y)
    +
    \mathbf{c}_x.
    \label{eq:proxy_normalization}
\end{equation}

We apply the same normalization \(\mathcal{N}_y\) to the proxy points and proxy anchors, denoting the normalized coordinates by \(\widetilde{\mathbf{y}}_i\) and \(\widetilde{\bar{\mathbf{y}}}_r\), respectively. Since \(\mathcal{N}_y\) consists only of translation and isotropic scaling, it preserves the canonical axes of the proxy.

We extract FCGF features~\cite{choy2019fully} from the input and the normalized proxy:
\begin{equation}
\begin{aligned}
    \mathcal{K}_{\mathcal{X}}
    =
    \left\{
        (\mathbf{x}_a,\mathbf{f}^{\mathcal{X}}_a)
    \right\}_{a=1}^{K_x},~~~ 
    \mathcal{K}_{\mathcal{Y}}
    =
    \left\{
        (\widetilde{\mathbf{y}}_c,
        \mathbf{f}^{\mathcal{Y}}_c)
    \right\}_{c=1}^{K_y}.
\end{aligned}
\label{eq:registration_features}
\end{equation}

For each input feature point, we identify its nearest input-side anchor and restrict the proxy candidates to the neighborhood of the paired proxy-side anchor:
\begin{equation}
\begin{aligned}
    \nu(a)
    &=
    \underset{1\leq r\leq N_a}
    {\operatorname{arg\,min}}\;
    \|\mathbf{x}_a-\bar{\mathbf{x}}_r\|_2,\\
    \mathcal{T}_a
    &=
    \left\{
        c\ \middle|\
        \|\widetilde{\mathbf{y}}_c
        -\widetilde{\bar{\mathbf{y}}}_{\nu(a)}\|_2
        < r_{\mathrm{gate}}
    \right\}.
\end{aligned}
\label{eq:anchor_neighborhood}
\end{equation}
Here, \(r_{\mathrm{gate}}\) is defined relative to the bounding-box diagonal of the normalized proxy feature cloud. Within \(\mathcal{T}_a\), the proxy point with the highest descriptor similarity is selected:
\begin{equation}
    c^\star(a)
    =
    \underset{c\in\mathcal{T}_a}
    {\operatorname{arg\,max}}\;
    \left\langle
        \mathbf{f}^{\mathcal{X}}_a,
        \mathbf{f}^{\mathcal{Y}}_c
    \right\rangle .
    \label{eq:anchor_constrained_match}
\end{equation}
Input feature points with empty candidate sets are discarded. The remaining correspondences form
\begin{equation}
    \mathcal{M}
    =
    \left\{
        (\mathbf{x}_a,
        \widetilde{\mathbf{y}}_{c^\star(a)})
        \ \middle|\
        \mathcal{T}_a\neq\varnothing
    \right\}.
    \label{eq:gated_matches}
\end{equation}
Thus, the anchors restrict the search regions, while the FCGF descriptors select the point correspondences.

We organize \(\mathcal{M}\) into a geometric compatibility graph based on pairwise distance preservation and enumerate its maximal cliques~\cite{zhang20233d}. Each clique \(\mathcal{Q}_{\ell}\) defines a geometrically coherent correspondence hypothesis, from which an input-to-proxy rigid transformation is estimated using the SVD-based solver~\cite{sorkine2017least}:
\begin{equation}
    (\widehat{\mathbf{R}}_{\ell},
     \widehat{\mathbf{t}}_{\ell})
    =
    \underset{
        \mathbf{R}\in\mathrm{SO}(3),\,
        \mathbf{t}\in\mathbb{R}^{3}
    }{\operatorname{arg\,min}}
    \sum_{
        (\mathbf{x},\widetilde{\mathbf{y}})
        \in\mathcal{Q}_{\ell}
    }
    \|
        \mathbf{R}\mathbf{x}
        +\mathbf{t}
        -\widetilde{\mathbf{y}}
    \|_2^2.
    \label{eq:pose_hypothesis}
\end{equation}

We select the hypothesis with the strongest geometric consensus over \(\mathcal{M}\) and refine it to obtain \((\widehat{\mathbf{R}},\widehat{\mathbf{t}})\). Since the proxy retains the target canonical axes, \(\widehat{\mathbf{R}}\) is the
desired canonicalizing rotation, whereas \(\widehat{\mathbf{t}}\) compensates only for coordinate-origin differences during registration. The final canonicalized shape is obtained by applying \(\widehat{\mathbf{R}}\) to the centered original input.

\begin{table*}[!t]
\centering
\caption{\label{tab:toys4k_big}
Quantitative comparison on 24 Toys4K categories using instance-level consistency (IC $\downarrow$) and category-level consistency (CC $\downarrow$); lower is better. The reported IC and CC values are multiplied by 10. The standard OrientAnything variants use a fixed rendering, whereas $+I^\dagger$ uses the view selected by CANIS and is included only as a diagnostic setting. Zero-shot denotes no category specific canonicalization training on Toys4K, and class-agnostic denotes one shared model and configuration across categories. Bold and underlined values indicate the best and second best results, respectively.
}
\footnotesize
\setlength{\tabcolsep}{2.9pt}
\renewcommand{\arraystretch}{1.10}
\resizebox{0.95\textwidth}{!}{%
\begin{tabular}{@{} l M{0.75cm} M{1.05cm} *{16}{c} @{}}
\toprule
\multirow{2}{*}{\textbf{Method}} & \multirow{2}{*}{\shortstack{\textbf{Zero-}\\[-1pt]\textbf{shot}}} & \multirow{2}{*}{\shortstack{\textbf{Class-}\\[-1pt]\textbf{agnostic}}}
& \multicolumn{2}{c}{\textbf{Airplane}}
& \multicolumn{2}{c}{\textbf{Bicycle}}
& \multicolumn{2}{c}{\textbf{Bus}}
& \multicolumn{2}{c}{\textbf{Bunny}}
& \multicolumn{2}{c}{\textbf{Boat}}
& \multicolumn{2}{c}{\textbf{Candy}}
& \multicolumn{2}{c}{\textbf{Car}}
& \multicolumn{2}{c}{\textbf{Crab}} \\
\cmidrule(lr){4-5}\cmidrule(lr){6-7}\cmidrule(lr){8-9}\cmidrule(lr){10-11}
\cmidrule(lr){12-13}\cmidrule(lr){14-15}\cmidrule(lr){16-17}\cmidrule(lr){18-19}
& & & \textbf{IC} & \textbf{CC} & \textbf{IC} & \textbf{CC} & \textbf{IC} & \textbf{CC} & \textbf{IC} & \textbf{CC} & \textbf{IC} & \textbf{CC} & \textbf{IC} & \textbf{CC} & \textbf{IC} & \textbf{CC} & \textbf{IC} & \textbf{CC} \\
\midrule
ShapeMatcher~\cite{di2024shapematcher} & $\times$ & $\times$
& $1.19$ & $1.51$ & $1.66$ & $1.38$ & $1.05$ & $1.38$ & $0.93$ & $1.25$ & $1.18$ & $1.54$ & $0.93$ & $2.19$ & $1.00$ & $0.98$ & $1.20$ & $1.54$ \\
CaCa~\cite{sun2021canonicalcapsules} & $\times$ & $\checkmark$
& $0.25$ & $0.46$ & $\underline{0.58}$ & $\underline{0.69}$
& $\mathbf{0.60}$ & $\mathbf{0.93}$
& $0.58$ & $0.91$ & $0.92$ & $0.97$ & $0.77$ & $2.07$ & $0.67$ & $0.82$ & $\underline{0.70}$ & $1.07$ \\
ConDor~\cite{sajnani2022_condor} & $\times$ & $\times$
& $\underline{0.14}$ & $0.53$ & $0.94$ & $0.92$ & $0.80$ & $1.51$ & $1.00$ & $1.35$ & $0.62$ & $1.26$ & $0.81$ & $\underline{2.01}$
& $\underline{0.52}$ & $\underline{0.77}$
& $0.94$ & $1.07$ \\
SR3D~\cite{scarvelissymmetry} & $\times$ & $\checkmark$
& $0.61$ & $0.90$ & $1.26$ & $1.79$ & $1.04$ & $1.30$ & $\underline{0.51}$ & $\underline{0.79}$
& $1.37$ & $1.60$ & $\mathbf{0.70}$ & $2.32$ & $0.59$ & $\mathbf{0.76}$
& $0.94$ & $0.97$ \\
\addlinespace[1.5pt]
OrientAnything~\cite{wangorient} & $\checkmark$ & $\checkmark$
& $0.32$ & $0.69$ & $1.74$ & $1.50$ & $1.14$ & $1.45$ & $0.87$ & $\underline{0.79}$
& $0.57$ & $0.82$ & $1.29$ & $2.23$ & $0.98$ & $1.01$ & $1.01$ & $\underline{0.92}$ \\
OrientAnythingV2~\cite{wang2026orient} & $\checkmark$ & $\checkmark$
& $0.21$ & $\underline{0.44}$
& $1.70$ & $1.32$ & $1.13$ & $1.60$ & $0.90$ & $0.85$ & $0.55$ & $0.76$ & $1.47$ & $2.29$ & $1.00$ & $0.91$ & $1.05$ & $0.93$ \\
OrientAnything + $I^\dagger$~\cite{wangorient} & $\checkmark$ & $\checkmark$
& $0.22$ & $0.53$ & $1.81$ & $1.75$ & $1.05$ & $1.44$ & $0.94$ & $0.90$ & $0.55$ & $\underline{0.72}$
& $1.16$ & $2.07$ & $0.90$ & $0.85$ & $0.88$ & $1.08$ \\
OrientAnythingV2 + $I^\dagger$~\cite{wang2026orient} & $\checkmark$ & $\checkmark$
& $0.47$ & $0.71$ & $1.67$ & $1.03$ & $1.02$ & $\underline{1.29}$
& $0.86$ & $0.97$ & $\underline{0.47}$ & $0.75$ & $1.11$ & $2.31$ & $0.79$ & $0.78$ & $0.97$ & $\mathbf{0.90}$ \\
\addlinespace[1.5pt]
\rowcolor{oursrow}
\textbf{Ours (CANIS)} & $\checkmark$ & $\checkmark$
& $\mathbf{0.10}$ & $\mathbf{0.42}$
& $\mathbf{0.37}$ & $\mathbf{0.31}$
& $\underline{0.71}$ & $1.31$ & $\mathbf{0.37}$ & $\mathbf{0.77}$
& $\mathbf{0.43}$ & $\mathbf{0.69}$
& $\underline{0.72}$ & $\mathbf{1.95}$
& $\mathbf{0.47}$ & $1.02$ & $\mathbf{0.53}$ & $\mathbf{0.90}$ \\
\midrule

\multirow{2}{*}{\textbf{Method}} & \multirow{2}{*}{\shortstack{\textbf{Zero-}\\[-1pt]\textbf{shot}}} & \multirow{2}{*}{\shortstack{\textbf{Class-}\\[-1pt]\textbf{agnostic}}}
& \multicolumn{2}{c}{\textbf{Dog}}
& \multicolumn{2}{c}{\textbf{Dragon}}
& \multicolumn{2}{c}{\textbf{Elephant}}
& \multicolumn{2}{c}{\textbf{Frog}}
& \multicolumn{2}{c}{\textbf{Giraffe}}
& \multicolumn{2}{c}{\textbf{Guitar}}
& \multicolumn{2}{c}{\textbf{Lizard}}
& \multicolumn{2}{c}{\textbf{Monkey}} \\
\cmidrule(lr){4-5}\cmidrule(lr){6-7}\cmidrule(lr){8-9}\cmidrule(lr){10-11}
\cmidrule(lr){12-13}\cmidrule(lr){14-15}\cmidrule(lr){16-17}\cmidrule(lr){18-19}
& & & \textbf{IC} & \textbf{CC} & \textbf{IC} & \textbf{CC} & \textbf{IC} & \textbf{CC} & \textbf{IC} & \textbf{CC} & \textbf{IC} & \textbf{CC} & \textbf{IC} & \textbf{CC} & \textbf{IC} & \textbf{CC} & \textbf{IC} & \textbf{CC} \\
\midrule
ShapeMatcher~\cite{di2024shapematcher} & $\times$ & $\times$
& $1.06$ & $1.27$ & $1.16$ & $1.47$ & $1.13$ & $1.40$ & $1.10$ & $1.44$ & $1.83$ & $1.02$ & $1.29$ & $1.97$ & $0.93$ & $1.10$ & $1.08$ & $1.29$ \\
CaCa~\cite{sun2021canonicalcapsules} & $\times$ & $\checkmark$
& $\underline{0.56}$ & $\underline{0.86}$
& $\underline{0.73}$ & $1.09$ & $\underline{0.57}$ & $1.04$ & $0.76$ & $1.25$ & $\underline{0.53}$ & $\underline{0.65}$
& $0.98$ & $1.51$ & $\underline{0.50}$ & $1.00$ & $\underline{0.68}$ & $1.04$ \\
ConDor~\cite{sajnani2022_condor} & $\times$ & $\times$
& $0.90$ & $1.10$ & $1.03$ & $1.29$ & $0.81$ & $1.16$ & $0.93$ & $1.31$ & $0.70$ & $0.75$ & $\mathbf{0.11}$ & $1.68$ & $0.74$ & $1.08$ & $0.85$ & $1.21$ \\
SR3D~\cite{scarvelissymmetry} & $\times$ & $\checkmark$
& $0.82$ & $1.12$ & $1.27$ & $1.51$ & $0.87$ & $1.05$ & $0.76$ & $0.92$ & $1.11$ & $1.58$ & $1.26$ & $1.52$ & $0.72$ & $\underline{0.85}$
& $0.95$ & $1.07$ \\
\addlinespace[1.5pt]
OrientAnything~\cite{wangorient} & $\checkmark$ & $\checkmark$
& $1.01$ & $1.02$ & $0.84$ & $\mathbf{0.83}$
& $0.60$ & $0.80$ & $0.79$ & $0.93$ & $1.04$ & $1.22$ & $1.28$ & $\mathbf{0.83}$
& $1.02$ & $1.06$ & $0.82$ & $0.88$ \\
OrientAnythingV2~\cite{wang2026orient} & $\checkmark$ & $\checkmark$
& $1.01$ & $1.18$ & $0.79$ & $0.89$ & $0.68$ & $0.82$ & $0.82$ & $\underline{0.91}$
& $0.91$ & $1.15$ & $1.35$ & $1.65$ & $1.01$ & $0.98$ & $0.78$ & $\underline{0.79}$ \\
OrientAnything + $I^\dagger$~\cite{wangorient} & $\checkmark$ & $\checkmark$
& $0.85$ & $1.02$ & $0.88$ & $0.93$ & $0.64$ & $0.79$ & $0.79$ & $0.93$ & $0.62$ & $0.77$ & $\underline{0.92}$ & $\underline{1.16}$
& $0.99$ & $1.07$ & $0.77$ & $0.87$ \\
OrientAnythingV2 + $I^\dagger$~\cite{wang2026orient} & $\checkmark$ & $\checkmark$
& $0.84$ & $0.91$ & $0.87$ & $0.89$ & $0.66$ & $\underline{0.69}$
& $\underline{0.71}$ & $0.93$ & $0.59$ & $0.79$ & $1.22$ & $1.25$ & $0.70$ & $0.95$ & $0.75$ & $\mathbf{0.76}$ \\
\addlinespace[1.5pt]
\rowcolor{oursrow}
\textbf{Ours (CANIS)} & $\checkmark$ & $\checkmark$
& $\mathbf{0.26}$ & $\mathbf{0.55}$
& $\mathbf{0.50}$ & $\underline{0.84}$
& $\mathbf{0.41}$ & $\mathbf{0.58}$
& $\mathbf{0.46}$ & $\mathbf{0.81}$
& $\mathbf{0.42}$ & $\mathbf{0.30}$
& $0.96$ & $1.25$ & $\mathbf{0.41}$ & $\mathbf{0.81}$
& $\mathbf{0.56}$ & $0.89$ \\
\midrule

\multirow{2}{*}{\textbf{Method}} & \multirow{2}{*}{\shortstack{\textbf{Zero-}\\[-1pt]\textbf{shot}}} & \multirow{2}{*}{\shortstack{\textbf{Class-}\\[-1pt]\textbf{agnostic}}}
& \multicolumn{2}{c}{\textbf{Mushroom}}
& \multicolumn{2}{c}{\textbf{Pencil}}
& \multicolumn{2}{c}{\textbf{Piano}}
& \multicolumn{2}{c}{\textbf{Robot}}
& \multicolumn{2}{c}{\textbf{Shoe}}
& \multicolumn{2}{c}{\textbf{Sofa}}
& \multicolumn{2}{c}{\textbf{Tree}}
& \multicolumn{2}{c}{\textbf{Truck}} \\
\cmidrule(lr){4-5}\cmidrule(lr){6-7}\cmidrule(lr){8-9}\cmidrule(lr){10-11}
\cmidrule(lr){12-13}\cmidrule(lr){14-15}\cmidrule(lr){16-17}\cmidrule(lr){18-19}
& & & \textbf{IC} & \textbf{CC} & \textbf{IC} & \textbf{CC} & \textbf{IC} & \textbf{CC} & \textbf{IC} & \textbf{CC} & \textbf{IC} & \textbf{CC} & \textbf{IC} & \textbf{CC} & \textbf{IC} & \textbf{CC} & \textbf{IC} & \textbf{CC} \\
\midrule
ShapeMatcher~\cite{di2024shapematcher} & $\times$ & $\times$
& $0.95$ & $1.59$ & $1.68$ & $0.97$ & $1.32$ & $1.36$ & $0.96$ & $1.18$ & $1.05$ & $0.98$ & $1.34$ & $1.23$ & $0.81$ & $1.06$ & $1.04$ & $1.34$ \\
CaCa~\cite{sun2021canonicalcapsules} & $\times$ & $\checkmark$
& $0.70$ & $1.60$ & $\underline{0.50}$ & $\underline{0.51}$
& $\mathbf{0.64}$ & $0.97$ & $\mathbf{0.70}$ & $1.01$ & $\mathbf{0.57}$ & $\underline{0.67}$
& $0.60$ & $\underline{0.63}$
& $0.56$ & $0.83$ & $0.76$ & $\underline{0.93}$ \\
ConDor~\cite{sajnani2022_condor} & $\times$ & $\times$
& $0.68$ & $1.53$ & $0.63$ & $0.57$ & $0.73$ & $0.97$ & $0.95$ & $1.39$ & $0.72$ & $1.01$ & $\underline{0.58}$ & $\mathbf{0.62}$
& $0.65$ & $0.88$ & $\mathbf{0.54}$ & $\mathbf{0.86}$ \\
SR3D~\cite{scarvelissymmetry} & $\times$ & $\checkmark$
& $0.68$ & $1.39$ & $1.84$ & $2.44$ & $\underline{0.68}$ & $\mathbf{0.93}$
& $\underline{0.73}$ & $\underline{0.60}$
& $0.75$ & $0.80$ & $\mathbf{0.53}$ & $0.70$ & $0.35$ & $0.72$ & $1.12$ & $1.38$ \\
\addlinespace[1.5pt]
OrientAnything~\cite{wangorient} & $\checkmark$ & $\checkmark$
& $0.69$ & $1.35$ & $2.20$ & $2.41$ & $0.89$ & $1.01$ & $1.13$ & $1.11$ & $0.91$ & $0.86$ & $1.33$ & $1.28$ & $0.31$ & $0.43$ & $1.02$ & $1.36$ \\
OrientAnythingV2~\cite{wang2026orient} & $\checkmark$ & $\checkmark$
& $0.69$ & $1.41$ & $2.03$ & $2.11$ & $0.72$ & $\underline{0.94}$
& $1.11$ & $1.08$ & $0.77$ & $0.91$ & $1.29$ & $1.25$ & $\underline{0.29}$ & $\mathbf{0.40}$
& $1.02$ & $1.24$ \\
OrientAnything + $I^\dagger$~\cite{wangorient} & $\checkmark$ & $\checkmark$
& $0.65$ & $\underline{1.22}$
& $2.44$ & $2.84$ & $0.83$ & $1.02$ & $1.00$ & $1.02$ & $0.82$ & $0.81$ & $1.24$ & $1.17$ & $0.31$ & $\underline{0.41}$
& $0.93$ & $1.15$ \\
OrientAnythingV2 + $I^\dagger$~\cite{wang2026orient} & $\checkmark$ & $\checkmark$
& $0.67$ & $\mathbf{1.21}$
& $2.12$ & $2.78$ & $1.02$ & $1.04$ & $0.94$ & $0.95$ & $0.83$ & $0.69$ & $1.13$ & $1.06$ & $\underline{0.29}$ & $\underline{0.41}$
& $0.96$ & $1.19$ \\
\addlinespace[1.5pt]
\rowcolor{oursrow}
\textbf{Ours (CANIS)} & $\checkmark$ & $\checkmark$
& $\mathbf{0.62}$ & $1.23$ & $\mathbf{0.30}$ & $\mathbf{0.03}$
& $0.69$ & $1.10$ & $\mathbf{0.70}$ & $\mathbf{0.59}$
& $\underline{0.59}$ & $\mathbf{0.66}$
& $0.91$ & $1.41$ & $\mathbf{0.28}$ & $0.42$ & $\underline{0.70}$ & $1.05$ \\
\bottomrule
\end{tabular}
}
\end{table*}

\section{Experiments}
\subsection{Experiment Setup}

We evaluate CANIS along seven aspects:
(1) \textbf{Comparative performance} against prior 3D canonicalization and image-based orientation methods;
(2) \textbf{Attribution} of performance gains to CANIS rather than the pretrained TRELLIS-OA backbone;
(3) \textbf{Ablations} on informative view selection, shape guidance, and semantic anchor-gated registration;
(4) \textbf{Robustness} to input rotations, partial observations, and real captured objects; 
(5) \textbf{Applicability} of semantic-anchor extraction, evaluating its effectiveness across different image-to-3D generators via attention-based cues; and
(6) \textbf{Efficiency} in terms of runtime and memory costs.
(7) \textbf{Downstream utility}, evaluating CANIS on 3D classification, part segmentation, and dense correspondence.
\vspace{0.5em}
\mypara{Datasets}
We evaluate on three datasets with complementary roles. Toys4K~\cite{stojanov2021using} is our primary external benchmark for comprehensive quantitative comparisons and component ablations over 24 categories. Objaverse-OA~\cite{lu2025orientation} provides a complementary benchmark whose object distribution is closer to the source domain of TRELLIS-OA; we use it for a focused comparison with representative 3D canonicalization methods and for qualitative visualization. OmniObject3D~\cite{wu2023omniobject3d} is used to qualitatively evaluate transfer to real captured objects.

\vspace{0.5em}
\mypara{Implementation Details}
We set the shape-guidance strength to $\alpha=0.10$, the semantic anchor gate radius to $r_{\mathrm{gate}}=0.2$, and use $8$ sparse-structure and $8$ SLAT sampling steps. 
For informative-view selection, all experiments use \(V=4\) candidate views. A larger \(V\) can improve view selection in practice. We order the cues as foreground area, normalized boundary length, within-foreground grayscale contrast, foreground color variation, foreground--background color separation, and boundary edge response. The corresponding weight vector is fixed to $\mathbf{w}=(1.0,\,0.15,\,0.20,\,0.15,\,0.30,\,0.10)^{\top}$ for all experiments.
We set \(\rho_j\geq\tau_{\mathrm{fg}}= 0.25\) and \(r_{gate}=0.2\) and $\Omega=\{4,\ldots,7\}\times\{10,\ldots,24\}$. We apply the same configuration to all evaluated categories without benchmark-specific training, fine-tuning, or test-time adaptation. The main quantitative experiments are conducted on NVIDIA RTX A6000 GPUs.

\vspace{0.5em}
\mypara{Evaluation Metrics}
CANIS uses the frozen TRELLIS-OA model~\cite{lu2025orientation} to define the target canonical frame. We therefore evaluate how consistently this frame is transferred under different input rotations and across object instances. Following ConDor~\cite{sajnani2022_condor}, we report IC and CC based on symmetric Chamfer distance. Qualitative results further illustrate the recovered upright and front-facing orientations. IC evaluates whether a \emph{single input point cloud} \(\mathcal{X}\) is canonicalized consistently under different input rotations.

Given a fixed set of random rotations \(\mathcal{R}=\{\mathbf{R}_k\}_{k=1}^{N_R}\subset\mathrm{SO}(3)\), where \(N_R=120\), we evaluate instance consistency by comparing the canonicalization results obtained before and after each rotation:
\vspace{-0.1cm}
\begin{equation}
    \mathrm{IC}(\mathcal{X})
    =
    \frac{1}{N_R}
    \sum_{k=1}^{N_R}
    \mathrm{CD}
    \left(
        \operatorname{can}(\mathbf{R}_k\mathcal{X}),
        \operatorname{can}(\mathcal{X})
    \right).
    \label{eq:instance_consistency}
\end{equation}
Here, \(\operatorname{can}(\cdot)\) denotes the canonicalized point cloud produced by the evaluated method. The reported IC is obtained by averaging \(\mathrm{IC}(\mathcal{X})\) over the test set. A lower IC indicates greater consistency under arbitrary input rotations.

Category consistency evaluates the agreement between independently canonicalized objects from the same category.  Let \(\mathcal{L}\) denote the set of test categories and \(\mathcal{D}_c\) the set of test point clouds belonging to category \(c\). For each \(\mathcal{X}_i\in\mathcal{D}_c\), we sample a fixed set \(\mathcal{N}_c(\mathcal{X}_i)\) of \(M=60\) other point clouds from the same category. Category consistency is defined as
\begin{equation}
    \mathrm{CC}
    =
    \frac{1}{|\mathcal{L}|}
    \sum_{c\in\mathcal{L}}
    \frac{1}{M|\mathcal{D}_c|}
    \sum_{\mathcal{X}_i\in\mathcal{D}_c}
    \sum_{\mathcal{X}_j\in\mathcal{N}_c(\mathcal{X}_i)}
    \mathrm{CD}
    \left(
        \operatorname{can}(\mathcal{X}_i),
        \operatorname{can}(\mathcal{X}_j)
    \right).
    \label{eq:category_consistency}
\end{equation}
For clarity, the reported IC and CC values in Table~\ref{tab:toys4k_big}--\ref{tab:ablation_20cls} are scaled by a factor of \textbf{10}, whereas the values those discussed in the text are reported on the original scale.

\begin{figure*}[t]
    \centering
    \vspace{\baselineskip}
    \includegraphics[width=0.95\linewidth]{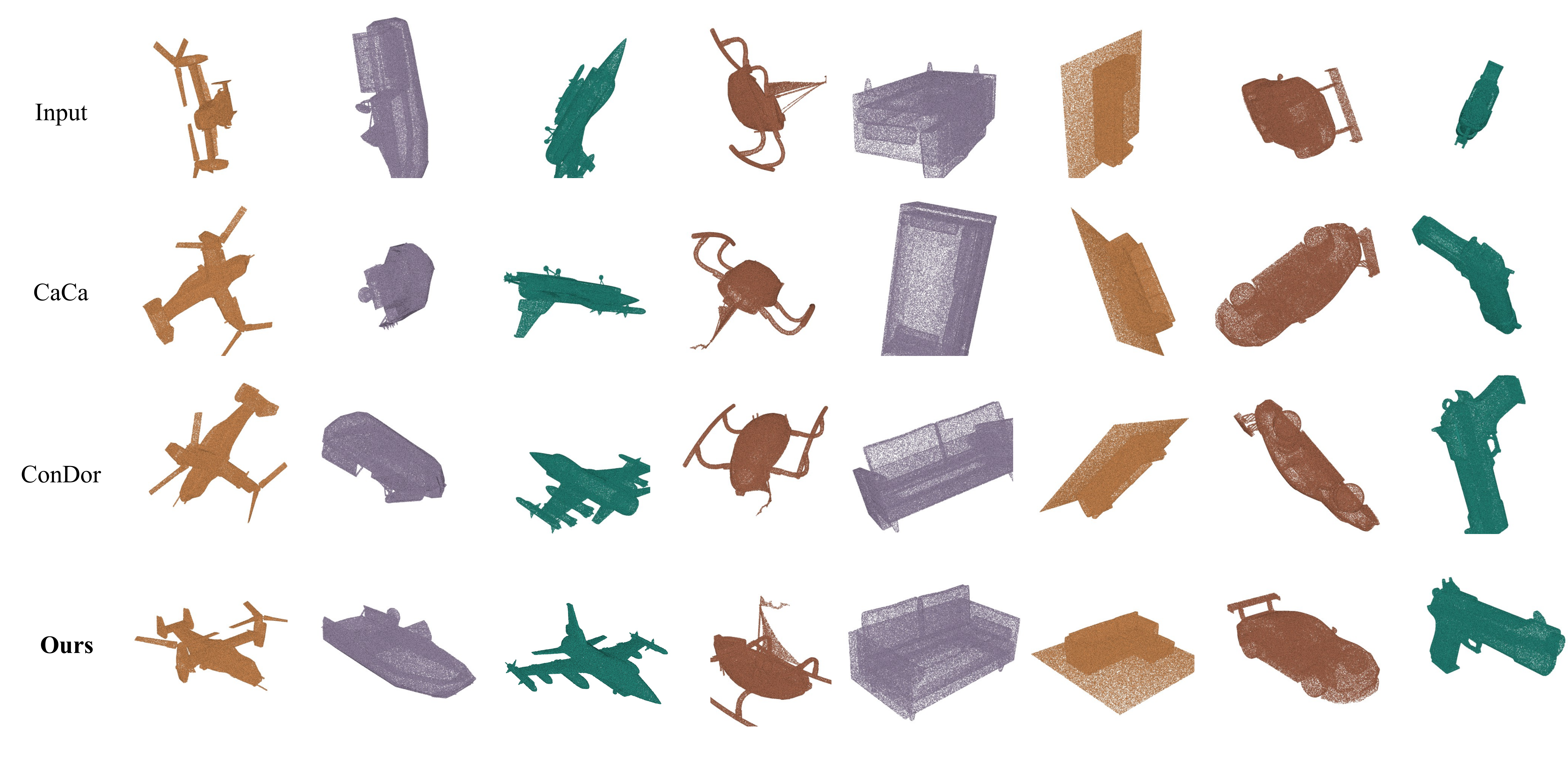}
    \vspace{-0.5cm}
    \caption{
    Qualitative comparison. Rows show the input and the canonicalization results of two representative baselines, CaCa and ConDor, together with those of CANIS. Columns show representative objects under arbitrary input rotations.
    }
   \vspace{-0.3cm}
    \label{fig:comparisons}
\end{figure*} 

\vspace{0.5em}
\mypara{Baselines}
We compare CANIS with four 3D canonicalization methods: CaCa~\cite{sun2021canonicalcapsules}, ConDor~\cite{sajnani2022_condor}, ShapeMatcher~\cite{di2024shapematcher}, and SR3D~\cite{scarvelissymmetry}. We also consider OneShot~\cite{jin2025one} and CanOverse~\cite{jin2026canoverse}, which address the same task but are excluded from quantitative comparison because no public implementations are available; OneShot additionally requires category-specific canonical templates. We also evaluate two image-based orientation estimators, OrientAnything~\cite{wangorient} and OrientAnythingV2~\cite{wang2026orient}. For these image-based methods, we compose the predicted object orientation with the known rendering-camera pose to obtain the canonicalizing rotation. The comprehensive Toys4K comparison includes all six methods, whereas the focused Objaverse-OA comparison uses CaCa and ConDor, for which directly compatible 3D canonicalization implementations and evaluation protocols are available. For the image-based baselines, we render each input object and compose the predicted object orientation with the corresponding camera pose to obtain the canonicalizing rotation. The standard OrientAnything and OrientAnythingV2 rows use a fixed rendering view. We additionally evaluate both methods using the informative view selected by CANIS; these rows are labeled `` + $I^\dagger$'' and serve as diagnostic variants rather than independent baseline methods. For a category-specific baseline without an exactly matched Toys4K checkpoint, we report the best result over all released checkpoints as an oracle upper bound that favors the baseline.

\begin{table}[t]
\caption{
Per category IC and CC results on 12 Objaverse OA categories. The reported values are multiplied by 10, and lower is better. The symbols ($\checkmark/\times$) below each method indicate zero-shot and class-agnostic support, respectively. Bold values indicate the best results, and the CANIS columns are shaded.}
\label{tab:objaverse_comparison}
\centering
\footnotesize
\setlength{\tabcolsep}{5pt}
\renewcommand{\arraystretch}{1.10}
\resizebox{0.42\textwidth}{!}{%
\begin{tabular}{@{}l
                cc
                cc
                >{\columncolor{oursrow}}c
                >{\columncolor{oursrow}}c@{}}
\toprule
\multirow{2}{*}{\textbf{Category}}
& \multicolumn{2}{c}{\shortstack{\textbf{CaCa}~\cite{sun2021canonicalcapsules}\\[-1pt]\scriptsize $\times/\checkmark$}}
& \multicolumn{2}{c}{\shortstack{\textbf{ConDor}~\cite{sajnani2022_condor}\\[-1pt]\scriptsize $\times/\times$}}
& \multicolumn{2}{>{\columncolor{oursrow}}c}{\shortstack{\textbf{CANIS (Ours)}\\[-1pt]\scriptsize $\checkmark/\checkmark$}}
\\
\cmidrule(lr){2-3}\cmidrule(lr){4-5}\cmidrule(lr){6-7}
& \textbf{IC} & \textbf{CC}
& \textbf{IC} & \textbf{CC}
& \textbf{IC} & \textbf{CC}
\\
\midrule
\textbf{Car}
& $0.65$ & $1.20$
& $0.34$ & $1.01$
& $\mathbf{0.21}$ & $\mathbf{0.89}$
\\
\textbf{Airplane}
& $0.36$ & $0.68$
& $0.42$ & $1.02$
& $\mathbf{0.10}$ & $\mathbf{0.46}$
\\
\textbf{Chair}
& $0.25$ & $0.69$
& $0.31$ & $0.66$
& $\mathbf{0.23}$ & $\mathbf{0.58}$
\\
\textbf{Couch}
& $1.18$ & $1.47$
& $0.53$ & $0.88$
& $\mathbf{0.31}$ & $\mathbf{0.42}$
\\
\textbf{Lamp}
& $0.74$ & $1.69$
& $0.53$ & $1.53$
& $\mathbf{0.32}$ & $\mathbf{1.12}$
\\
\textbf{Watercraft}
& $0.63$ & $1.08$
& $0.41$ & $0.94$
& $\mathbf{0.37}$ & $\mathbf{0.78}$
\\
\textbf{Table}
& $0.93$ & $1.26$
& $0.64$ & $1.07$
& $\mathbf{0.47}$ & $\mathbf{1.00}$
\\
\textbf{Speaker}
& $0.72$ & $2.31$
& $\mathbf{0.57}$ & $1.97$
& $0.65$ & $\mathbf{1.72}$
\\
\textbf{Cabinet}
& $1.37$ & $1.90$
& $0.45$ & $1.33$
& $\mathbf{0.34}$ & $\mathbf{1.10}$
\\
\textbf{Firearm}
& $0.65$ & $1.00$
& $0.48$ & $0.99$
& $\mathbf{0.18}$ & $\mathbf{0.49}$
\\
\textbf{Monitor}
& $1.54$ & $1.20$
& $0.83$ & $1.29$
& $\mathbf{0.42}$ & $\mathbf{0.82}$
\\
\textbf{Cellphone}
& $0.79$ & $1.57$
& $0.55$ & $1.32$
& $\mathbf{0.23}$ & $\mathbf{0.95}$
\\
\bottomrule
\end{tabular}
}
\vspace{-0.2cm}
\end{table}

\subsection{Comparison with Prior Methods}
\vspace{0.5em}
\mypara{Comprehensive Comparison on Toys4K} 
We conduct a comprehensive quantitative evaluation on 24 categories of the external Toys4K benchmark. We compare CANIS with four 3D canonicalization methods: CaCa~\cite{sun2021canonicalcapsules}, ConDor~\cite{sajnani2022_condor}, ShapeMatcher~\cite{di2024shapematcher}, and SR3D~\cite{scarvelissymmetry}, as well as two image-based orientation estimators, OrientAnything~\cite{wangorient} and OrientAnythingV2~\cite{wang2026orient}. All methods are evaluated on the same test objects using the same IC and CC definitions. For the image-based methods, we compose the predicted object orientation with the corresponding rendering-camera pose to obtain the canonicalizing rotation. Their standard configurations use a fixed rendering view. We additionally report variants that replace the fixed rendering with the informative view selected by CANIS, allowing us to assess whether the proposed view selection strategy also benefits external orientation estimators. Because these variants incorporate a component of CANIS, we treat them as controlled diagnostic configurations rather than independent prior methods. CANIS obtains the lowest average IC and CC over the 24 categories. Based on the per-category values in Table~\ref{tab:toys4k_big}, CANIS achieves average IC/CC scores of 0.052/0.083, compared with 0.064/0.098 for the strongest independent baseline, CaCa. When all reported settings, including the two variants using our view selector, are considered, CANIS ranks first in 31 and within the top two in 36 of the 48 category metric entries. 

\begin{table*}[!t]
\caption{Ablation studies on Toys4K~\cite{stojanov2021using}. We compare the TRELLIS-OA baselines, direct matching with FPFH or FCGF, shape guidance, and semantic anchor gating. The reported IC and CC values are multiplied by 10, and lower is better. Bold values indicate the best results.}
\label{tab:ablation_20cls}
\centering
\footnotesize
\setlength{\tabcolsep}{3.2pt}
\renewcommand{\arraystretch}{1.10}
\resizebox{0.93\textwidth}{!}{%
\begin{tabular}{@{}l *{16}{c}@{}}
\toprule
\multirow{2}{*}{\textbf{Method}}
& \multicolumn{2}{c}{\textbf{Airplane}}
& \multicolumn{2}{c}{\textbf{Bicycle}}
& \multicolumn{2}{c}{\textbf{Bus}}
& \multicolumn{2}{c}{\textbf{Bunny}}
& \multicolumn{2}{c}{\textbf{Boat}}
& \multicolumn{2}{c}{\textbf{Candy}}
& \multicolumn{2}{c}{\textbf{Car}}
& \multicolumn{2}{c}{\textbf{Crab}} \\
\cmidrule(lr){2-3}\cmidrule(lr){4-5}\cmidrule(lr){6-7}\cmidrule(lr){8-9}
\cmidrule(lr){10-11}\cmidrule(lr){12-13}\cmidrule(lr){14-15}\cmidrule(lr){16-17}
& \textbf{IC} & \textbf{CC} & \textbf{IC} & \textbf{CC} & \textbf{IC} & \textbf{CC} & \textbf{IC} & \textbf{CC} & \textbf{IC} & \textbf{CC} & \textbf{IC} & \textbf{CC} & \textbf{IC} & \textbf{CC} & \textbf{IC} & \textbf{CC} \\
\midrule

Plain TRELLIS-OA & $0.30$ & $0.90$ & $1.00$ & $0.90$ & $2.20$ & $2.90$ & $0.90$ & $2.10$ & $1.20$ & $1.60$ & $2.20$ & $5.30$ & $1.10$ & $3.00$ & $1.30$ & $2.40$ \\
Plain TRELLIS-OA(Multi-view) & $0.20$ & $1.00$ & $0.90$ & $0.70$ & $1.60$ & $2.80$ & $0.90$ & $1.70$ & $1.10$ & $1.50$ & $1.70$ & $3.80$ & $1.20$ & $2.00$ & $1.20$ & $1.80$ \\

\addlinespace[1pt]
Direct FPFH Matching    & $0.15$ & $0.72$ & $0.69$ & $0.53$ & $1.24$ & $2.25$ & $0.58$ & $1.23$ & $0.77$ & $1.16$ & $1.11$ & $3.08$ & $0.79$ & $1.63$ & $0.93$ & $1.56$ \\
Direct FCGF Matching     & $0.17$ & $0.72$ & $0.60$ & $0.55$ & $1.17$ & $2.10$ & $0.59$ & $1.20$ & $0.71$ & $1.07$ & $1.07$ & $2.92$ & $0.67$ & $1.51$ & $0.81$ & $1.43$ \\
w/o Shape Guidance    & $0.14$ & $0.52$ & $0.47$ & $0.38$ & $0.86$ & $1.67$ & $0.50$ & $1.01$ & $0.53$ & $0.94$ & $0.92$ & $2.51$ & $0.69$ & $1.40$ & $0.69$ & $1.17$ \\
\addlinespace[1pt]
\rowcolor{oursrow}
Semantic Anchors + FPFH          & $0.23$ & $0.57$ & $0.50$ & $0.45$ & $0.99$ & $1.58$ & $\mathbf{0.31}$ & $0.99$ & $0.67$ & $0.85$ & $0.81$ & $2.15$ & $0.65$ & $1.29$ & $0.75$ & $1.11$ \\
\rowcolor{oursrow}
Semantic Anchors + FCGF (\textbf{Ours})               & $\mathbf{0.10}$ & $\mathbf{0.42}$ & $\mathbf{0.37}$ & $\mathbf{0.31}$ & $\mathbf{0.71}$ & $\mathbf{1.31}$ & $0.37$ & $\mathbf{0.77}$ & $\mathbf{0.43}$ & $\mathbf{0.69}$ & $\mathbf{0.72}$ & $\mathbf{1.95}$ & $\mathbf{0.47}$ & $\mathbf{1.02}$ & $\mathbf{0.53}$ & $\mathbf{0.90}$ \\
\midrule

\multirow{2}{*}{\textbf{Method}}
& \multicolumn{2}{c}{\textbf{Dog}}
& \multicolumn{2}{c}{\textbf{Dragon}}
& \multicolumn{2}{c}{\textbf{Elephant}}
& \multicolumn{2}{c}{\textbf{Frog}}
& \multicolumn{2}{c}{\textbf{Giraffe}}
& \multicolumn{2}{c}{\textbf{Guitar}}
& \multicolumn{2}{c}{\textbf{Lizard}}
& \multicolumn{2}{c}{\textbf{Monkey}} \\
\cmidrule(lr){2-3}\cmidrule(lr){4-5}\cmidrule(lr){6-7}\cmidrule(lr){8-9}
\cmidrule(lr){10-11}\cmidrule(lr){12-13}\cmidrule(lr){14-15}\cmidrule(lr){16-17}
& \textbf{IC} & \textbf{CC} & \textbf{IC} & \textbf{CC} & \textbf{IC} & \textbf{CC} & \textbf{IC} & \textbf{CC} & \textbf{IC} & \textbf{CC} & \textbf{IC} & \textbf{CC} & \textbf{IC} & \textbf{CC} & \textbf{IC} & \textbf{CC} \\
\midrule

Plain TRELLIS-OA & $0.70$ & $1.30$ & $1.20$ & $2.30$ & $1.00$ & $1.30$ & $1.30$ & $2.20$ & $1.30$ & $0.70$ & $2.60$ & $3.20$ & $1.20$ & $2.40$ & $1.80$ & $2.20$ \\
Plain TRELLIS-OA(Multi-view) & $0.60$ & $1.20$ & $1.00$ & $2.00$ & $1.00$ & $1.20$ & $1.00$ & $1.90$ & $1.10$ & $0.60$ & $2.50$ & $2.50$ & $0.90$ & $1.50$ & $1.50$ & $1.80$ \\
\addlinespace[1pt]
Direct FPFH Matching    & $0.48$ & $0.90$ & $0.78$ & $1.34$ & $0.64$ & $0.89$ & $0.74$ & $1.23$ & $0.63$ & $0.40$ & $1.49$ & $2.02$ & $0.67$ & $1.26$ & $0.80$ & $1.33$ \\
Direct FCGF Matching     & $0.49$ & $0.87$ & $0.77$ & $1.26$ & $0.57$ & $0.89$ & $0.67$ & $1.15$ & $0.60$ & $0.43$ & $1.41$ & $1.87$ & $0.60$ & $1.24$ & $0.84$ & $1.23$ \\
w/o Shape Guidance    & $0.34$ & $0.68$ & $0.67$ & $1.14$ & $0.52$ & $0.78$ & $0.62$ & $1.04$ & $0.57$ & $0.40$ & $1.29$ & $1.64$ & $0.48$ & $1.02$ & $0.68$ & $1.20$ \\
\addlinespace[1pt]
\rowcolor{oursrow}
Semantic Anchors + FPFH         & $0.50$ & $0.77$ & $0.69$ & $0.99$ & $\mathbf{0.38}$ & $0.75$ & $0.60$ & $1.06$ & $0.65$ & $\mathbf{0.27}$ & $1.19$ & $1.44$ & $0.65$ & $1.00$ & $0.74$ & $1.07$ \\
\rowcolor{oursrow}
Semantic Anchors + FCGF (\textbf{Ours})                 & $\mathbf{0.26}$ & $\mathbf{0.55}$ & $\mathbf{0.50}$ & $\mathbf{0.84}$ & $0.41$ & $\mathbf{0.58}$ & $\mathbf{0.46}$ & $\mathbf{0.81}$ & $\mathbf{0.42}$ & $0.30$ & $\mathbf{0.96}$ & $\mathbf{1.25}$ & $\mathbf{0.41}$ & $\mathbf{0.81}$ & $\mathbf{0.56}$ & $\mathbf{0.89}$ \\
\midrule

\multirow{2}{*}{\textbf{Method}}
& \multicolumn{2}{c}{\textbf{Mushroom}}
& \multicolumn{2}{c}{\textbf{Pencil}}
& \multicolumn{2}{c}{\textbf{Piano}}
& \multicolumn{2}{c}{\textbf{Robot}}
& \multicolumn{2}{c}{\textbf{Shoe}}
& \multicolumn{2}{c}{\textbf{Sofa}}
& \multicolumn{2}{c}{\textbf{Tree}}
& \multicolumn{2}{c}{\textbf{Truck}}\\

\cmidrule(lr){2-3}\cmidrule(lr){4-5}\cmidrule(lr){6-7}\cmidrule(lr){8-9}
\cmidrule(lr){10-11}\cmidrule(lr){12-13}\cmidrule(lr){14-15}\cmidrule(lr){16-17}
& \textbf{IC} & \textbf{CC} & \textbf{IC} & \textbf{CC} & \textbf{IC} & \textbf{CC} & \textbf{IC} & \textbf{CC} & \textbf{IC} & \textbf{CC} & \textbf{IC} & \textbf{CC} & \textbf{IC} & \textbf{CC} & \textbf{IC} & \textbf{CC}\\
\midrule

Plain TRELLIS-OA & $1.70$ & $2.80$ & $0.90$ & $0.10$ & $1.90$ & $3.30$ & $1.70$ & $1.60$ & $1.90$ & $1.60$ & $2.40$ & $3.90$ & $0.80$ & $1.20$ & $2.00$ & $2.60$ \\
Plain TRELLIS-OA(Multi-view) & $1.50$ & $2.30$ & $0.80$ & $0.10$ & $1.80$ & $2.50$ & $1.60$ & $1.20$ & $1.60$ & $1.50$ & $2.40$ & $3.40$ & $0.60$ & $0.80$ & $1.80$ & $2.40$ \\
\addlinespace[1pt]
Direct FPFH Matching    & $0.93$ & $2.12$ & $0.52$ & $0.07$ & $1.16$ & $1.74$ & $1.13$ & $0.96$ & $0.91$ & $1.03$ & $1.42$ & $2.22$ & $0.46$ & $0.72$ & $1.19$ & $1.82$ \\
Direct FCGF Matching     & $0.84$ & $2.03$ & $0.46$ & $0.08$ & $1.17$ & $1.62$ & $0.99$ & $0.91$ & $0.84$ & $0.94$ & $1.42$ & $2.13$ & $0.45$ & $0.70$ & $1.18$ & $1.72$ \\
w/o Shape Guidance    & $0.79$ & $1.66$ & $0.38$ & $\mathbf{0.03}$ & $0.91$ & $1.43$ & $0.98$ & $0.72$ & $0.75$ & $0.86$ & $1.27$ & $1.85$ & $0.34$ & $0.54$ & $0.97$ & $1.44$ \\
\addlinespace[1pt]
\rowcolor{oursrow}
Semantic Anchors + FPFH        & $\mathbf{0.61}$ & $\mathbf{1.16}$ & $0.42$ & $0.04$ & $0.98$ & $1.35$ & $\mathbf{0.61}$ & $0.80$ & $0.74$ & $0.89$ & $1.22$ & $1.57$ & $0.46$ & $0.62$ & $0.90$ & $1.29$ \\
\rowcolor{oursrow}
Semantic Anchors + FCGF (\textbf{Ours})       & $0.62$ & $1.23$ & $\mathbf{0.30}$ & $\mathbf{0.03}$ & $\mathbf{0.69}$ & $\mathbf{1.10}$ & $0.70$ & $\mathbf{0.59}$ & $\mathbf{0.59}$ & $\mathbf{0.66}$ & $\mathbf{0.91}$ & $\mathbf{1.41}$ & $\mathbf{0.28}$ & $\mathbf{0.42}$ & $\mathbf{0.70}$ & $\mathbf{1.05}$ \\
\bottomrule
\end{tabular}
}
\vspace{-0.2cm}
\end{table*}

\vspace{0.5em}
\mypara{Complementary Comparison on Objaverse-OA}
We complement the external Toys4K evaluation with a focused experiment on Objaverse-OA, whose object distribution is closer to the source domain of the TRELLIS-OA backbone. We compare with CaCa~\cite{sun2021canonicalcapsules} and ConDor~ \cite{sajnani2022_condor}, two representative 3D canonicalization methods with publicly available implementations and directly compatible IC/CC evaluation protocols. This experiment supports a controlled comparison and the qualitative analysis in Fig.~\ref{fig:comparisons}; broader baseline coverage is provided by the Toys4K experiment.
Table~\ref{tab:objaverse_comparison} reports IC and CC over 12 Objaverse-OA categories. CANIS obtains the lowest value in 23 of the 24 category--metric entries; the only exception is IC on Speaker, for which ConDor performs better. Averaged over the 12 categories, CANIS obtains IC/CC scores of 0.032/0.086, compared with 0.051/0.117 for ConDor and 0.082/0.134 for CaCa. These results indicate that the improvements observed on the external Toys4K benchmark also persist in this Objaverse-based evaluation.

\vspace{0.5em}
\mypara{Qualitative Results} 
Fig.~\ref{fig:comparisons} compares three methods on representative Objaverse-OA examples. Across the displayed input rotations, CANIS more consistently recovers upright and front-facing orientations, whereas the representative baselines, CaCa and ConDor, exhibit larger orientation variations in several challenging cases. These qualitative observations align with the IC and CC results.

\subsection{Backbone and Component Analysis}
We conduct controlled ablations on the same 24-category Toys4K benchmark to disentangle the contributions of CANIS from the
orientation prior encoded by its pretrained TRELLIS-OA backbone. The analysis addresses two questions. First, we compare the complete CANIS pipeline with plain TRELLIS-OA under fixed-view and multi-view conditioning to examine whether the backbone alone accounts for the observed canonicalization performance. Second, we evaluate the principal design choices of CANIS by removing semantic anchor gating and disabling shape guidance. All variants are evaluated on the same test objects using identical preprocessing and IC/CC protocols, with the remaining settings held fixed whenever applicable. Table~\ref{tab:ablation_20cls} reports per category IC and CC on the 24 Toys4K categories, with the corresponding averages discussed below; lower values are better.

\vspace{0.5em}
\mypara{Contribution beyond the Pretrained Backbone}
As shown in Table~\ref{tab:ablation_20cls}, plain TRELLIS-OA obtains average IC/CC scores of 0.144/0.216. Providing multiple input views improves the scores to 0.127/0.176, but both backbone-only settings remain substantially worse than the full CANIS pipeline (0.052/0.083). This indicates that the final performance cannot be explained solely by directly using the generator's pose prior. We observe the same trend in proxy fidelity: the mean CD-L2 decreases from 0.1579 with fixed-view TRELLIS-OA and 0.1455 with multi-view TRELLIS-OA to 0.1210 with informative view selection and shape guidance. 

\begin{figure}[t]
    \centering
    \vspace{\baselineskip}
    \includegraphics[width=\linewidth]{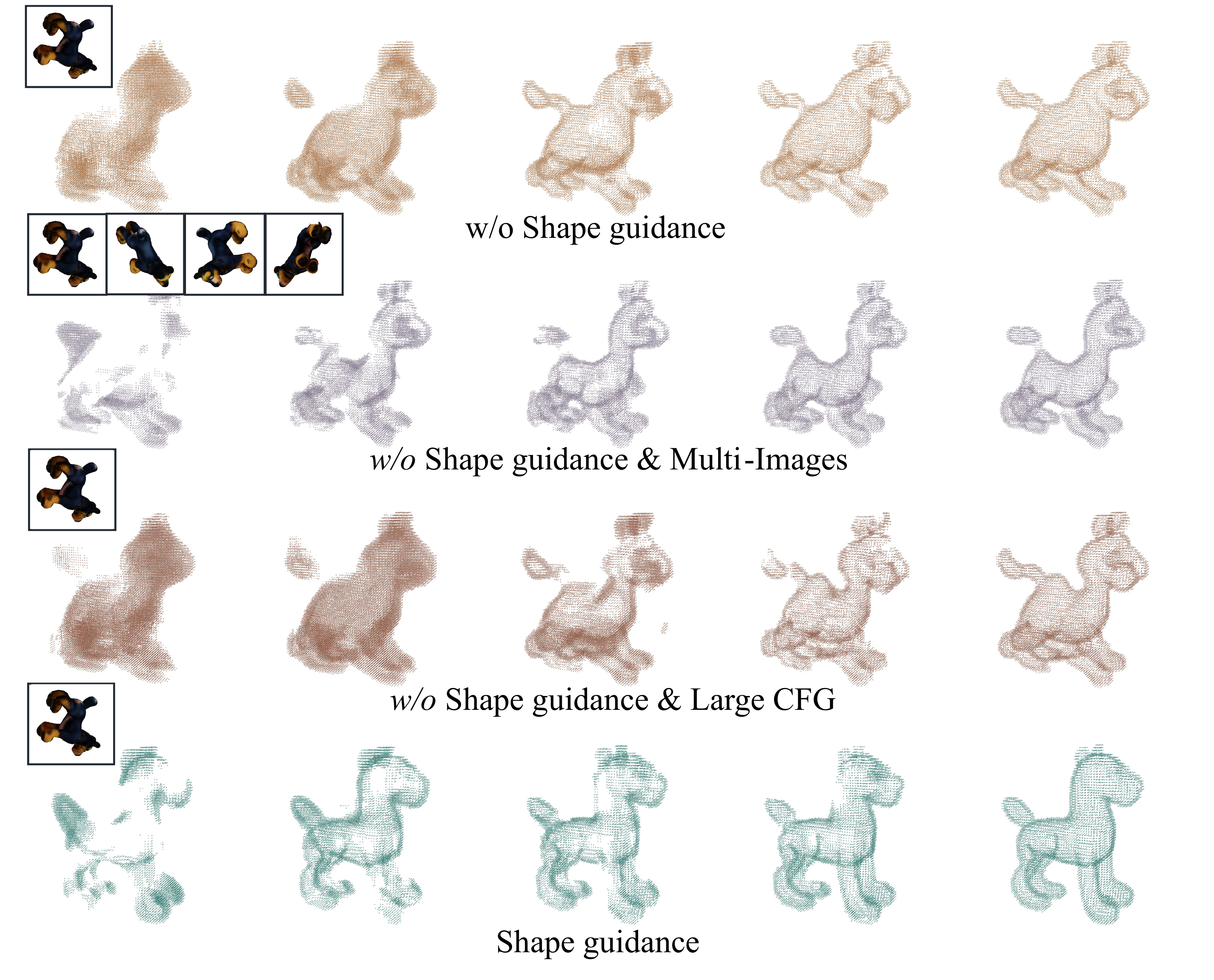}
    \vspace{-0.8cm}
   \caption{
   Effect of shape guidance on proxy synthesis under challenging views. Rows compare generation without shape guidance, with multiview prompting, with increased classifier free guidance (CFG), and with the proposed shape guidance. Shape guidance better preserves the global structure of the input in these examples.
   }
    \label{fig:shape_guidance}
    \vspace{-0.3cm}
\end{figure}

\vspace{0.5em}
\mypara{Effect of Semantic Anchors}
Removing semantic anchors degrades both metrics for both local geometric descriptors. With FPFH, semantic anchor gating reduces the average IC/CC from approximately 0.084/0.134 to 0.068/0.100. With FCGF, the full model improves over direct matching from approximately 0.080/0.127 to 0.052/0.083. These paired comparisons show that semantic anchor gating improves downstream rotation and category consistency for both descriptor choices. The consistent gains across FPFH and FCGF indicate that the improvement does not depend on a particular descriptor. By restricting candidate matches to semantically aligned neighborhoods, the anchors suppress geometrically plausible but semantically incorrect correspondences on repeated or approximately symmetric parts.

\vspace{0.5em}
\mypara{Effect of Shape Guidance}
Removing shape guidance increases average IC/CC from
0.052/0.083 to approximately 0.068/0.108. The full model
performs better in 47 of the 48 category metric entries, with one tie. Fig.~\ref{fig:shape_guidance} provides complementary qualitative evidence: under the displayed atypical views, increasing image guidance or adding views alone can still produce distorted proxy geometry, whereas shape guidance better preserves the global structure of the input.

\begin{table*}[t]
\centering

\begin{minipage}[t]{0.485\textwidth}
\vspace{0pt}
\centering
\captionof{table}{
Effect of selected view rank on proxy fidelity using a 1,000 object subset of Toys4K. We report the mean, median, and standard deviation of Chamfer distance (CD). Lower mean and median values indicate more faithful proxies. Bold values indicate the best results, and the Top 1 column is shaded.}

\label{tab:toys4k_view_rank_ablation}
\small
\setlength{\tabcolsep}{2.5pt}
\renewcommand{\arraystretch}{1.250}

\resizebox{0.96\linewidth}{!}{%
\begin{tabular}{@{}l >{\columncolor{oursrow}}c c c c c@{}}
\toprule
\textbf{Metric}
& \textbf{Top-1}
& \textbf{Top-2}
& \textbf{Top-6}
& \textbf{Top-8}
& \textbf{Top-10}
\\
\midrule
CD mean $\downarrow$
& $\mathbf{0.1098}$
& $0.1105$
& $0.1314$
& $0.1266$
& $0.1379$
\\
CD median $\downarrow$
& $\mathbf{0.0968}$
& $0.0994$
& $0.1107$
& $0.1017$
& $0.1200$
\\
CD standard deviation
& $0.0590$
& $0.0603$
& $0.0591$
& $0.0740$
& $0.0580$
\\
\bottomrule
\end{tabular}%
}
\end{minipage}
\hfill
\begin{minipage}[t]{0.49\textwidth}
\vspace{0pt}
\centering
\captionof{table}{
Effect of shape guidance strength $\alpha$ on proxy fidelity for Toys4K. Mean and median Chamfer distance (CD) are reported, and lower is better. The proper range $0.08\leq\alpha\leq0.12$ is shaded, and $\alpha=0.10$ is used in the other experiments. Bold values indicate the best results.
}
\label{tab:alpha_cd_l2}
\vspace{-0.01cm}

\footnotesize
\setlength{\tabcolsep}{2.4pt}
\renewcommand{\arraystretch}{1.10}

\resizebox{0.95\linewidth}{!}{%
\begin{tabular}{@{}l c c
    >{\columncolor{oursrow}}c
    >{\columncolor{oursrow}}c
    >{\columncolor{oursrow}}c c@{}}
\toprule
\multirow{2}{*}{\textbf{Metric}}
& \multicolumn{6}{c}{\textbf{Guidance strength weight} $\alpha$}
\\
\cmidrule(lr){2-7}
& $\mathbf{0.00}$
& $\mathbf{0.05}$
& $\mathbf{0.08}$
& $\mathbf{0.10}$
& $\mathbf{0.12}$
& $\mathbf{0.15}$
\\
\midrule
CD mean $\downarrow$
& $0.1356$
& $0.1253$
& $\mathbf{0.1206}$
& $0.1210$
& $0.1212$
& $0.1411$
\\
CD median $\downarrow$
& $0.1169$
& $0.1110$
& $0.0999$
& $0.1014$
& $\mathbf{0.0961}$
& $0.0984$
\\
\bottomrule
\end{tabular}%
}
\end{minipage}
\vspace{-0.4cm}
\end{table*}

\begin{figure*}[!t]
    \centering
    \vspace{\baselineskip}
    \includegraphics[width=0.95\linewidth]{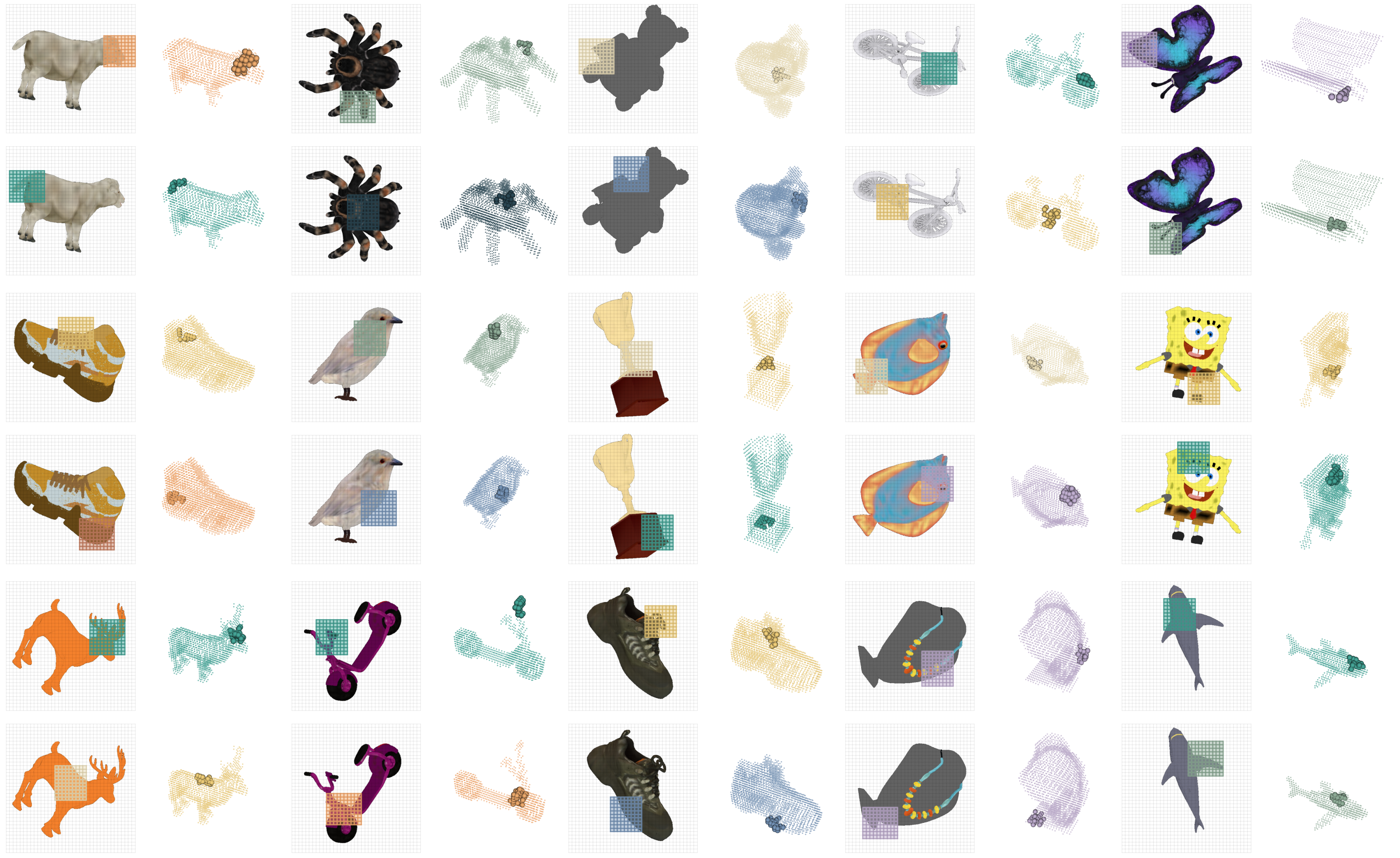}
    \caption{\label{fig:2d_3d} Patch--to--cluster associations used by PCAR. Each example shows a selected image block in the rendered input and the proxy voxel cluster associated with that block by aggregated cross attention. Matching colors identify paired image and proxy regions.
    }
    \vspace{-0.3cm}
  \label{fig:2d_3d}
\end{figure*}

\begin{figure*}[t]
    \centering
    \includegraphics[width=0.95\linewidth]{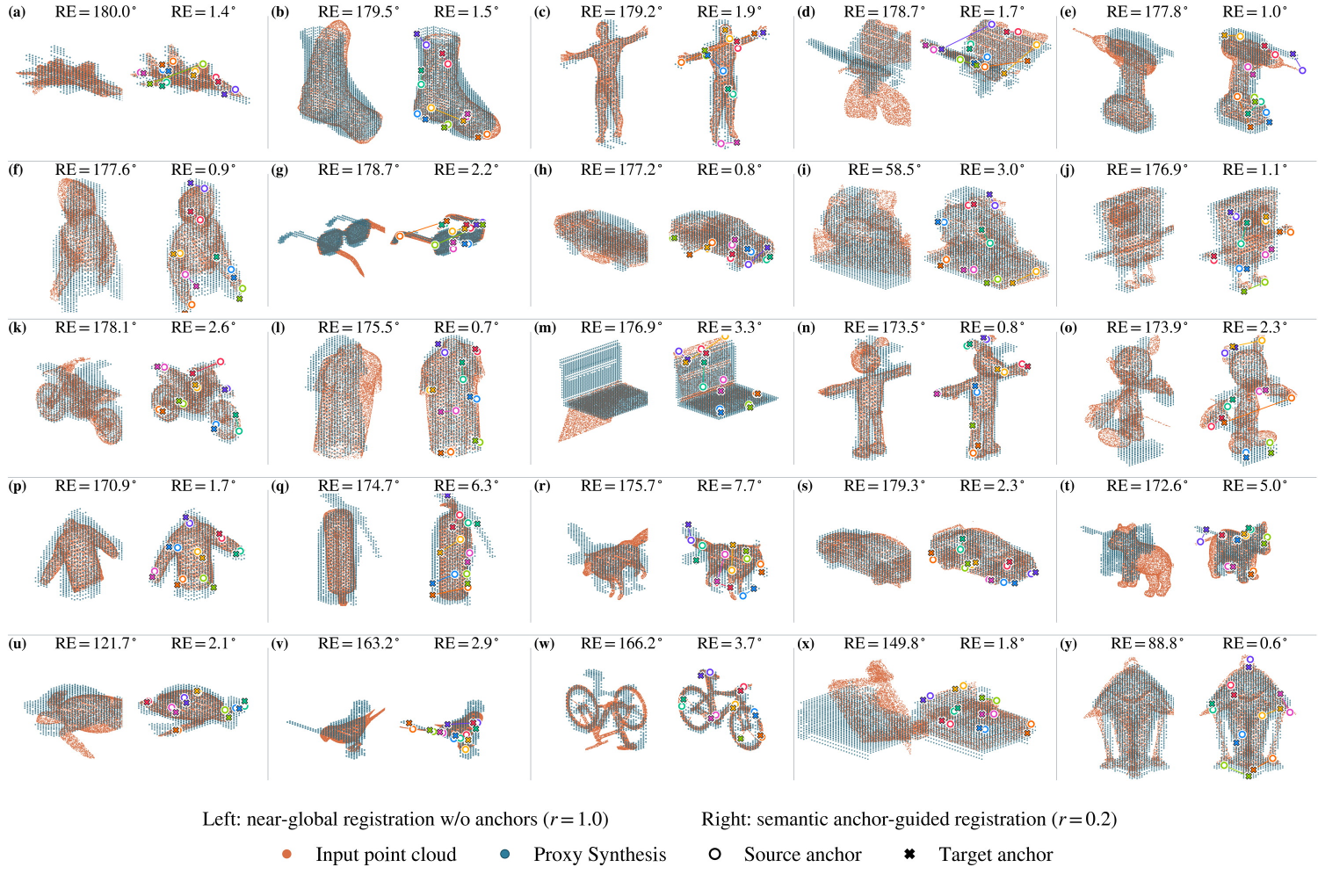}
    \vspace{-0.5cm}
    \caption{Visualizations of semantic anchor-guided point-cloud registration. We highlight examples that appear geometrically reasonable under near-global registration but are semantically incorrect (e.g., mis-oriented front/back or up/down). Each pair compares near global FCGF registration using $r_{\mathrm{gate}}=1.0$ (left) with anchor constrained registration using $r_{\mathrm{gate}}=0.2$ (right). Dense and sparse points denote the input point cloud and canonical proxy, respectively. Circles and crosses indicate source and target anchors, respectively. RE denotes rotation error in degrees; lower is better.}
   \label{fig:anchor_registration}
   \vspace{-0.55cm}
\end{figure*}

\begin{figure*}[!htb]
    \centering
    \vspace{\baselineskip}
    \includegraphics[width=0.95\linewidth]{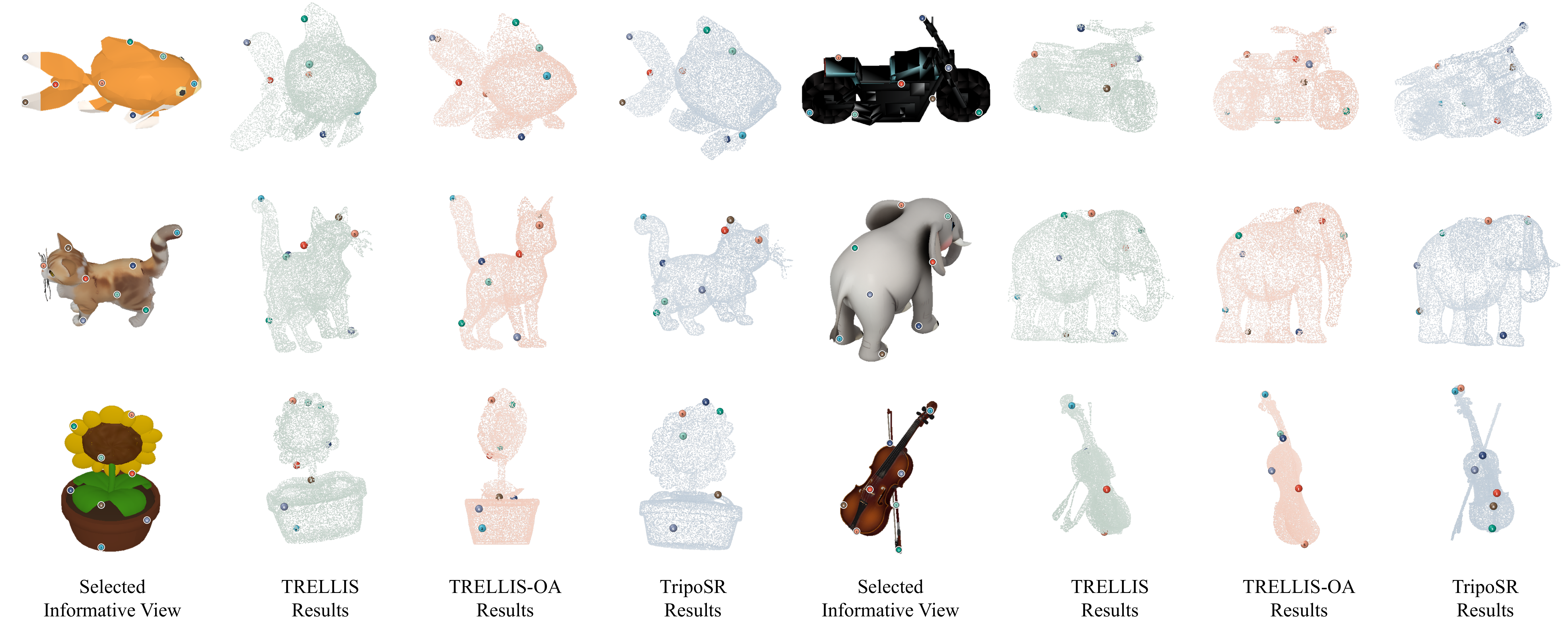}
    \vspace{-0.2cm}
   \caption{
   Attention--induced correspondences across three image to 3D generators. Each four column group shows the selected informative view followed by outputs from TRELLIS~\cite{xiang2025structured}, TRELLIS-OA~\cite{lu2025orientation}, and TripoSR~\cite{tochilkin2024triposr}. Markers of the same color indicate 3D responses associated with the same image location. For visualization, the TRELLIS and TripoSR outputs are coarsely aligned with the TRELLIS-OA output using these correspondences.}
    \label{fig:cross_generator_correspondence}
   \vspace{-0.4cm}
\end{figure*}

\subsection{Analysis of SCPS}
We further analyze the two key design choices in SCPS: informative view selection and shape-guided proxy synthesis. We first examine whether the view-ranking function identifies renderings that yield more faithful canonical proxies and whether the selected views also benefit external image-based orientation estimators. We then vary the shape-guidance strength $\alpha$ to characterize its effect on proxy fidelity. We report proxy CD-L2 for analyses of the intermediate canonical proxy and IC/CC for downstream canonicalization performance. These experiments complement the component ablations by directly examining the behavior of the SCPS stage.

\vspace{0.5em}
\mypara{Informative View Ranking}
We evaluate the view-ranking stage on a 1K-sample subset of Toys4K using proxy CD-L2. As shown in Table~\ref{tab:toys4k_view_rank_ablation}, the top-ranked view obtains the lowest mean and median CD among the reported ranks, with values of 0.1098 and 0.0968, respectively. Although the results are not strictly monotonic across ranks, the Top-1 view yields the lowest sample mean and median CD among the evaluated ranks. The small difference between Top-1 and Top-2 should be interpreted without a significance claim.

\vspace{0.5em}
\mypara{Transferability of the Selected View}
Replacing the fixed rendering with the selected informative view improves the average performance of both image-based orientation estimators. As shown in Table~\ref{tab:toys4k_big}, the average IC/CC changes from 0.099/0.112 to 0.092/0.111 for OrientAnything and it changes from 0.097/0.112 to 0.090/0.105 for OrientAnythingV2. The benefit is not uniform across all categories, but the aggregate improvement indicates that the view selector captures cues useful beyond the TRELLIS-OA proxy generator.

\vspace{0.5em}
\mypara{Shape Guidance Strength}
Table~\ref{tab:alpha_cd_l2} evaluates the guidance weight $\alpha$. The lowest mean and median CD are obtained at $\alpha=0.08$ and $\alpha=0.12$, respectively. We use $\alpha=0.10$, for which the reference weight decreases from $0.10$ to $0.0125$ over eight sampling steps, preserving early structural cues while leaving later generation image-dominated. Increasing $\alpha$ to $0.15$ degrades the mean CD, suggesting that excessive guidance can introduce image-inconsistent structure.

\subsection{Analysis of PCAR}

We analyze three components of PCAR: the radius used for semantic anchor gating, the cross-generator consistency of attention-induced correspondences, and the choice of local geometric descriptor (FPFH or FCGF) for candidate correspondence extraction. These experiments examine how each component affects correspondence disambiguation and downstream canonicalization.

\vspace{0.5em}
\mypara{Semantic Gating Radius}
The gate radius \(r_{\mathrm{gate}}\) controls the spatial extent of candidate correspondences around each proxy-side semantic anchor. At the scale used in our experiments, setting \(r_{\mathrm{gate}}=1.0\) makes the correspondence search effectively near-global and approximates direct FCGF matching. Reducing the radius to \(r_{\mathrm{gate}}=0.2\) restricts each source keypoint to target keypoints near its paired proxy-side anchor. On Toys4K, this semantic restriction reduces the average IC/CC from approximately 0.080/0.127 to 0.052/0.083.
Fig.~\ref{fig:anchor_registration} further illustrates the difference. Near-global FCGF matching can produce geometrically plausible registrations with incorrect semantic orientations, particularly front--back and up--down flips. Restricting the correspondence search to paired anchor neighborhoods suppresses matches between geometrically similar but semantically different parts and recovers the correct orientation in the displayed cases. Both settings use the same FCGF descriptor and pose estimator, isolating the effect of semantic anchor gating.

\mypara{Qualitative Cross-Generator Correspondence Consistency}
We examine whether the attention aggregation in PCAR produces semantically consistent correspondences across different image-to-3D generators. As shown in Fig.~\ref{fig:cross_generator_correspondence}, the same informative view selected by SCPS is used to condition TRELLIS~\cite{xiang2025structured}, TRELLIS-OA~\cite{lu2025orientation}, and TripoSR~\cite{tochilkin2024triposr}. Markers of the same color associate a selected 2D image location with its attention-aggregated 3D response in each generated output. We pair these 3D locations through their common image regions and estimate rigid rotations that coarsely align the TRELLIS and TripoSR outputs with the TRELLIS-OA output, which defines the reference canonical frame. Despite differences in generated geometry and initial orientation, the markers remain associated with corresponding semantic parts in the displayed examples. These examples show that attention-induced regions can remain semantically related across the three tested generators. 

\vspace{0.5em}
\mypara{Correspondence Features}
With semantic anchors enabled, replacing FPFH with FCGF reduces the average IC/CC from approximately 0.068/0.100 to 0.052/0.083. FCGF performs better in 42 of the 48 category--metric entries, including 20 of 24 IC entries and 22 of 24 CC entries. We therefore use FCGF in the full model.

\begin{figure*}[t]
    \centering
    \subfloat[Partially missing data.\label{fig:partial}]{%
        \includegraphics[width=0.475\textwidth]{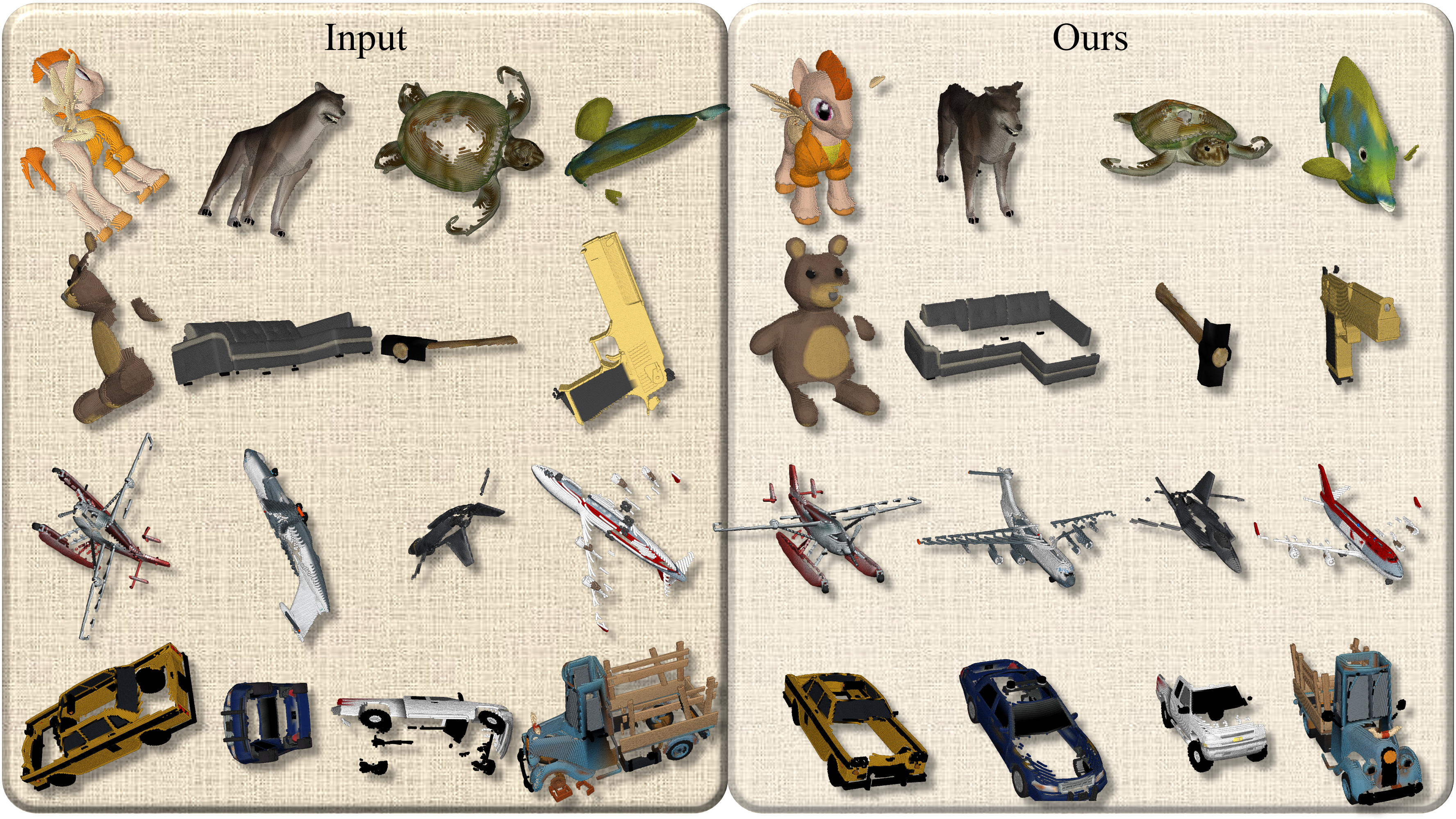}%
    }%
    \hspace{0.15cm}
    \subfloat[Real-world OmniObject3D data.\label{fig:omni3d_real}]{%
        \includegraphics[width=0.475\textwidth]{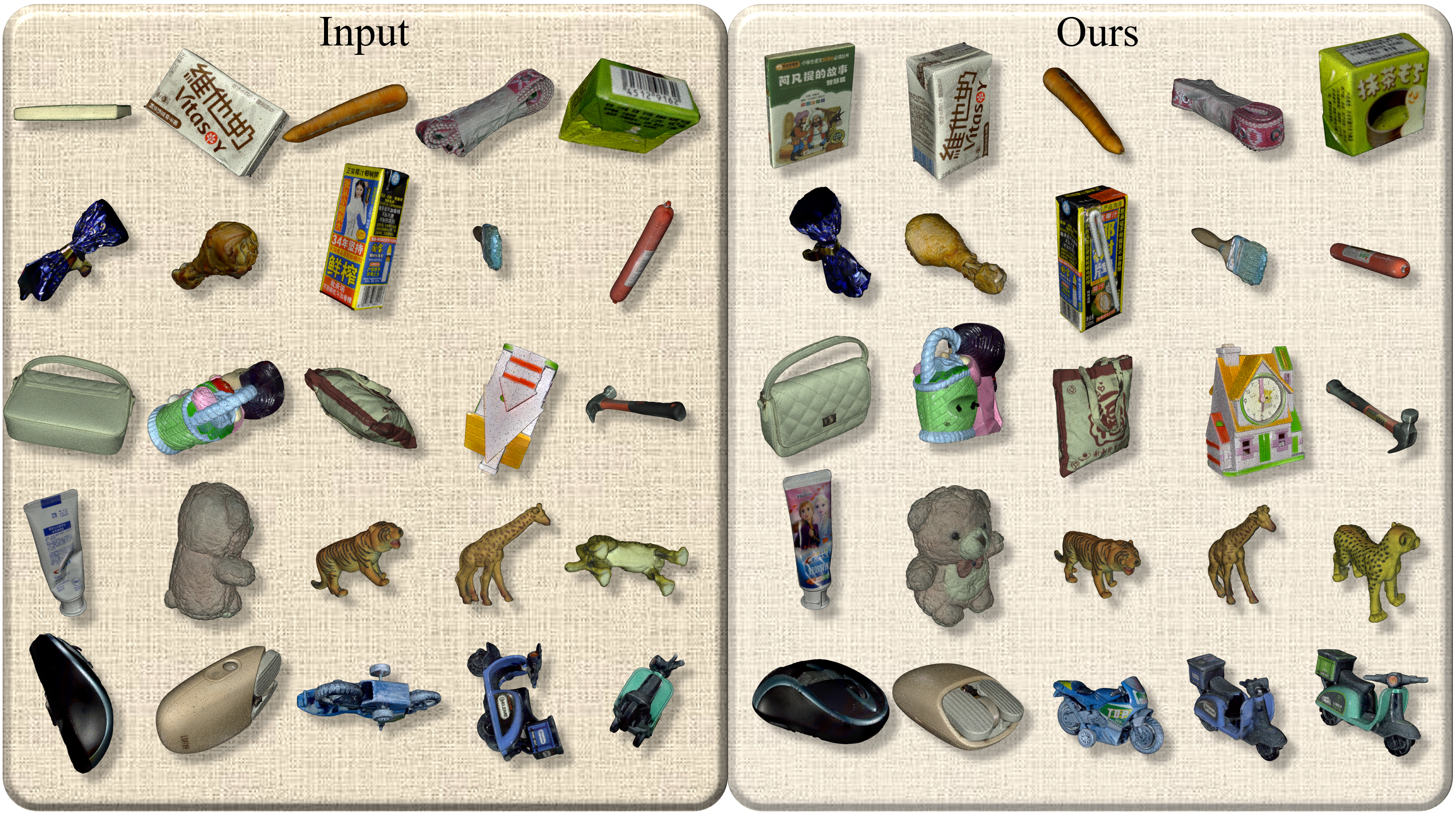}%
    }
    \caption{
    Canonicalization of partial observations and real captures. (a) Partially observed inputs and the corresponding CANIS results. (b) OmniObject3D scans and the corresponding CANIS results. Within each panel, the inputs are shown on the left and the results on the right.}
    \label{fig:partial_real_combined}
    \vspace{-0.2cm}
\end{figure*}

\begin{figure*}[t]
    \centering
    \includegraphics[width=\linewidth]
        {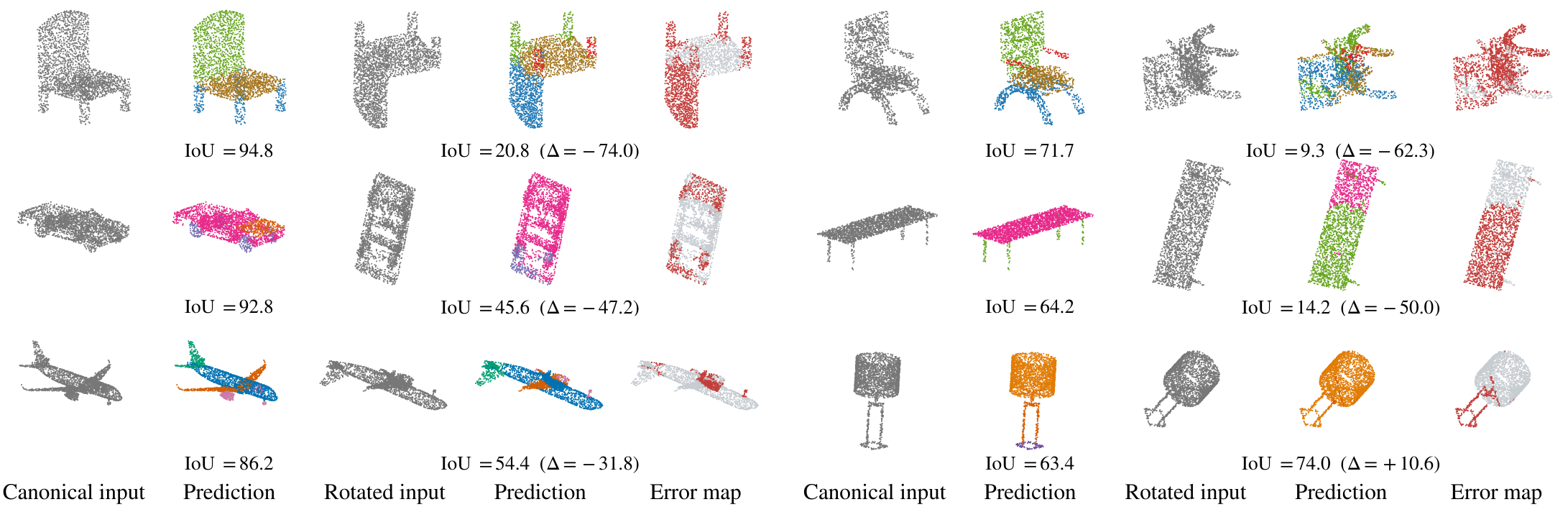}
    \caption{
    Effect of CANIS canonicalization on Point MAE~\cite{pointmae} part segmentation on ShapeNetPart~\cite{shapenetpart} subset. Each example shows, from left to right, the CANIS canonicalized input and its prediction, the same rotated input before canonicalization and its prediction, and the error map for the latter. IoU is reported below each prediction, and $\Delta=\mathrm{IoU}_{\mathrm{rot}}-\mathrm{IoU}_{\mathrm{can}}$. Red points indicate labels that differ from the ground truth.}
    \label{fig:segmentation}
    \vspace{-0.3cm}
\end{figure*}

\vspace{-0.2cm}
\subsection{Robustness and Real-World Transfer}

\mypara{Partial Observations}
To assess sensitivity to incomplete geometry, we construct partial input point clouds with varying degrees of incompleteness by fusing a limited set of available observations. Lower observation coverage corresponds to more severe missingness. Fig.~\ref{fig:partial} presents representative results under the evaluated incompleteness levels.

\vspace{0.2em}
\mypara{Transfer to Real Captures}
Finally, we apply CANIS to real captured objects from OmniObject3D. As shown in Fig.~\ref{fig:omni3d_real}, CANIS produces plausible upright and front-facing orientations for the displayed scans despite being evaluated primarily on synthetic objects. These qualitative examples suggest promising transfer to captured data.

\begin{figure*}
    \centering
    \vspace{\baselineskip}
    \includegraphics[width=\linewidth]{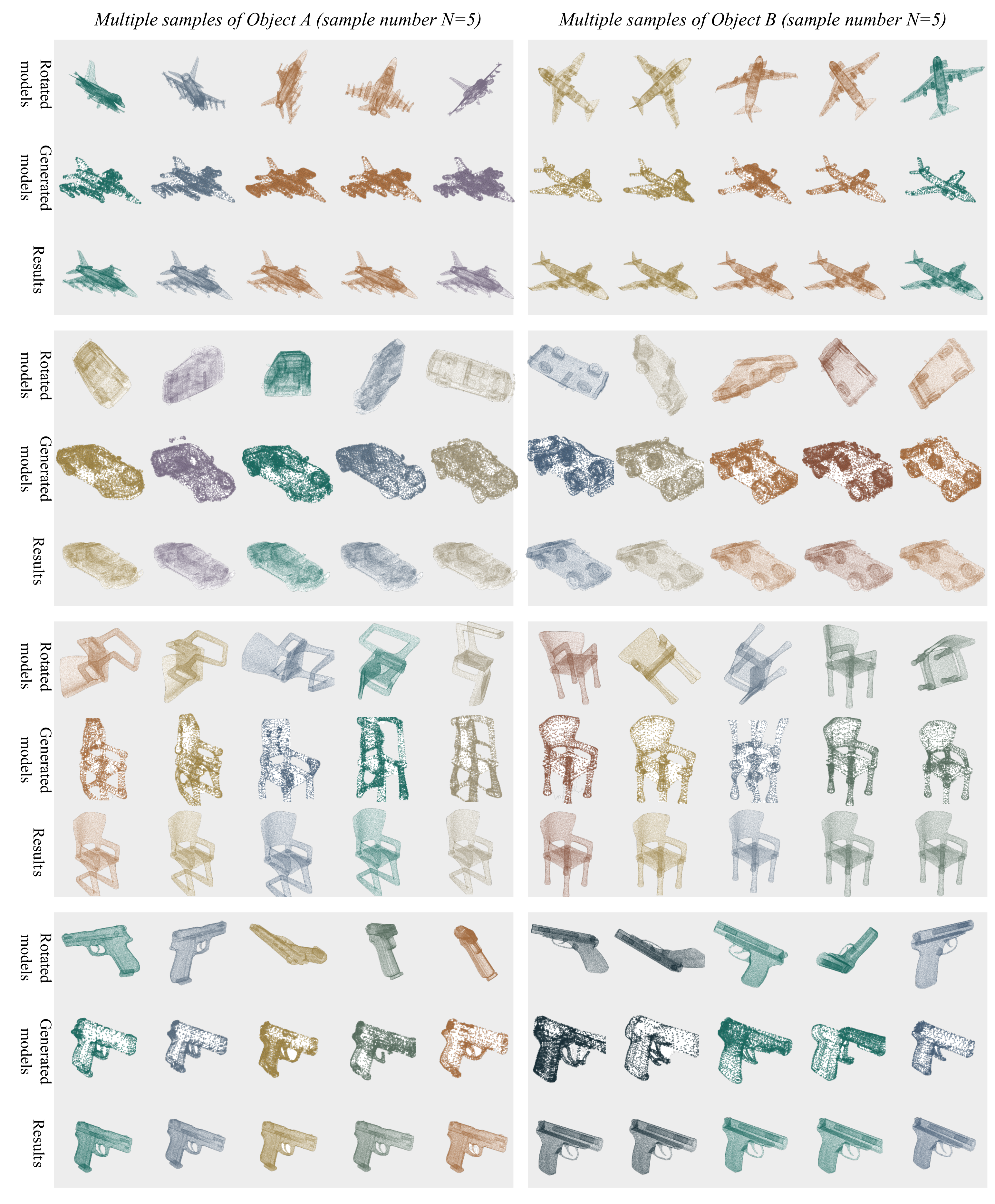}
    \vspace{-1cm}
   \caption{
   Consistency under input rotations. For each object, the first row shows five randomly rotated inputs, the second row shows the corresponding generated canonical proxies, and the third row shows the final canonicalized inputs. Similar final orientations across columns illustrate the consistency measured by IC.}
    \label{fig:multi_sample}
\end{figure*}

\vspace{0.5em}
\mypara{Initial Rotations}
To complement the quantitative IC evaluation, Fig.~\ref{fig:multi_sample} visualizes the outputs obtained from five rotated versions of each example. CANIS produces similar canonical orientations across the displayed rotations, providing a qualitative illustration of the consistency measured by IC.

\vspace{-0.2cm}
\subsection{Applications}
\vspace{0.4em}

\mypara{3D Part Segmentation}
We further evaluate CANIS for downstream part segmentation on a subset of ShapeNetPart~\cite{shapenetpart} using Point-MAE~\cite{pointmae} and PointNeXt~\cite{qian2022pointnext}. The same \(\mathrm{SO}(3)\)-rotated inputs are segmented directly or after CANIS canonicalization. As reported in Table~\ref{tab:shapenetpart_downstream}, CANIS increases instance mIoU from 31.43\% to 85.67\% for Point-MAE and from 35.27\% to 86.23\% for PointNeXt, with consistent gains across the remaining metrics. Fig.~\ref{fig:segmentation} further shows reduced semantic-part confusion. Under this evaluation protocol, CANIS recovers most of the performance lost after applying SO(3) rotations.

\begin{table*}[t]
  \centering
  \caption{Part-segmentation performance on the evaluated ShapeNetPart subset~\cite{shapenetpart}. The same SO(3) rotated inputs are evaluated directly (Rotated) and after CANIS canonicalization (Canonical). All metrics are reported as percentages, and higher values are better. The better result for each backbone and metric is shown in bold.
  }
  \label{tab:shapenetpart_downstream}

  \footnotesize
  \setlength{\tabcolsep}{3.5pt}
  \renewcommand{\arraystretch}{1.12}

  \begin{tabular}{@{}ll ccc ccc ccc@{}}
    \toprule
    & &
    \multicolumn{3}{c}{IoU Metrics}
    & \multicolumn{3}{c}{Accuracy and Dice}
    & \multicolumn{3}{c}{Macro Metrics} \\
    \cmidrule(lr){3-5}
    \cmidrule(lr){6-8}
    \cmidrule(lr){9-11}

    Method
    & Input
    & \shortstack{Instance\\mIoU}
    & \shortstack{Class\\mIoU}
    & \shortstack{Global Part\\mIoU}
    & \shortstack{Instance\\mDice}
    & \shortstack{Point\\Accuracy}
    & \shortstack{Mean Part\\Accuracy}
    & Precision
    & Recall
    & F1 \\
    \midrule

    \rowcolor{oursrow}
    Point-MAE~\cite{pointmae}
    & Canonical
    & $\mathbf{85.67}$
    & $\mathbf{82.31}$
    & $\mathbf{82.28}$
    & $\mathbf{89.39}$
    & $\mathbf{94.60}$
    & $\mathbf{88.91}$
    & $\mathbf{90.79}$
    & $\mathbf{88.91}$
    & $\mathbf{89.49}$ \\

    & Rotated
    & $31.43$
    & $34.66$
    & $27.21$
    & $38.76$
    & $51.18$
    & $37.78$
    & $39.51$
    & $37.78$
    & $36.53$ \\
    \midrule

    \rowcolor{oursrow}
    PointNeXt~\cite{qian2022pointnext}
    & Canonical
    & $\mathbf{86.23}$
    & $\mathbf{81.50}$
    & $\mathbf{80.53}$
    & $\mathbf{89.98}$
    & $\mathbf{94.52}$
    & $\mathbf{87.05}$
    & $\mathbf{91.25}$
    & $\mathbf{87.05}$
    & $\mathbf{88.40}$ \\

    & Rotated
    & $35.27$
    & $37.63$
    & $27.89$
    & $42.58$
    & $52.34$
    & $39.09$
    & $40.57$
    & $39.09$
    & $37.39$ \\

    \bottomrule
  \end{tabular}
\end{table*}

\begin{table*}[t]
\centering
\footnotesize
\caption{
ModelNet40 classification using the original inputs, the corresponding CANIS canonicalized inputs, and the same SO(3) rotated inputs. OA and mAcc are reported as percentages. $\Delta$OA is the signed difference from the original input in percentage points and is computed before rounding. Bold and underlined values indicate the best and second best OA and mAcc results, respectively.
}
\label{tab:classification}
\setlength{\tabcolsep}{3.3pt}
\renewcommand{\arraystretch}{1.12}

\newcolumntype{C}{>{\centering\arraybackslash}p{1.05cm}}

\begin{tabular}{
  @{}
  l
  >{\centering\arraybackslash}p{1.10cm}
  *{8}{C}
  @{}
}
  \toprule
  & &
  \multicolumn{2}{c}{Original}
  & \multicolumn{3}{c}{
      \shortstack[c]{
        Canonical\\[-1pt]
        {\scriptsize (Rotated + CANIS)}
      }
    }
  & \multicolumn{3}{c}{
      \shortstack[c]{
        Rotated\\[-1pt]
        {\scriptsize ($\mathrm{SO}(3)$)}
      }
    } \\
  \cmidrule(lr){3-4}
  \cmidrule(lr){5-7}
  \cmidrule(lr){8-10}

  Method
  & \shortstack{Params.\\(M)}
  & OA & mAcc
  & OA & mAcc & $\Delta$OA
  & OA & mAcc & $\Delta$OA \\
  \midrule

  PointNeXt~\cite{qian2022pointnext}
  & $4.52$
  & $\mathbf{93.96}$
  & $\underline{91.14}$
  & $\mathbf{93.23}$
  & $\mathbf{90.55}$
  & $-0.73$
  & $\underline{11.95}$
  & $11.51$
  & $-82.01$ \\

  Point-MAE~\cite{pointmae}
  & $22.10$
  & $92.46$
  & $89.42$
  & $91.98$
  & $88.83$
  & $-0.49$
  & $11.55$
  & $\underline{11.54}$
  & $-80.92$ \\

  PointGPT-L~\cite{chen2023pointgpt}
  & $360.46$
  & $\underline{93.40}$
  & $\mathbf{91.56}$
  & $\underline{92.54}$
  & $\underline{90.52}$
  & $-0.85$
  & $\mathbf{15.52}$
  & $\mathbf{15.65}$
  & $-77.88$ \\

  \bottomrule
\end{tabular}

\end{table*}

\vspace{0.5em}
\mypara{3D Object Classification}
We evaluate CANIS as a preprocessing step for downstream classification on ModelNet40. For each of the 2,468 test shapes, we sample a fixed rotation from \(\mathrm{SO}(3)\) and evaluate each pretrained classifier on the unmodified input, the rotated input, and the same input after CANIS canonicalization. All classifiers use 1,024 points per object without adaptation. We report overall accuracy (OA) and mean class accuracy (mAcc). Table~\ref{tab:classification} reports an OA decrease of at least 77.88 percentage points under arbitrary rotations. Canonicalization with CANIS yields OA values above 91.98\%, with deviations from the original accuracies limited to 0.85 percentage points. The corresponding recovery in mAcc further demonstrates the effectiveness of CANIS canonicalization for downstream classification under arbitrary input rotations.

\begin{figure*}[!t]
    \centering
    \includegraphics[width=0.98\linewidth]
        {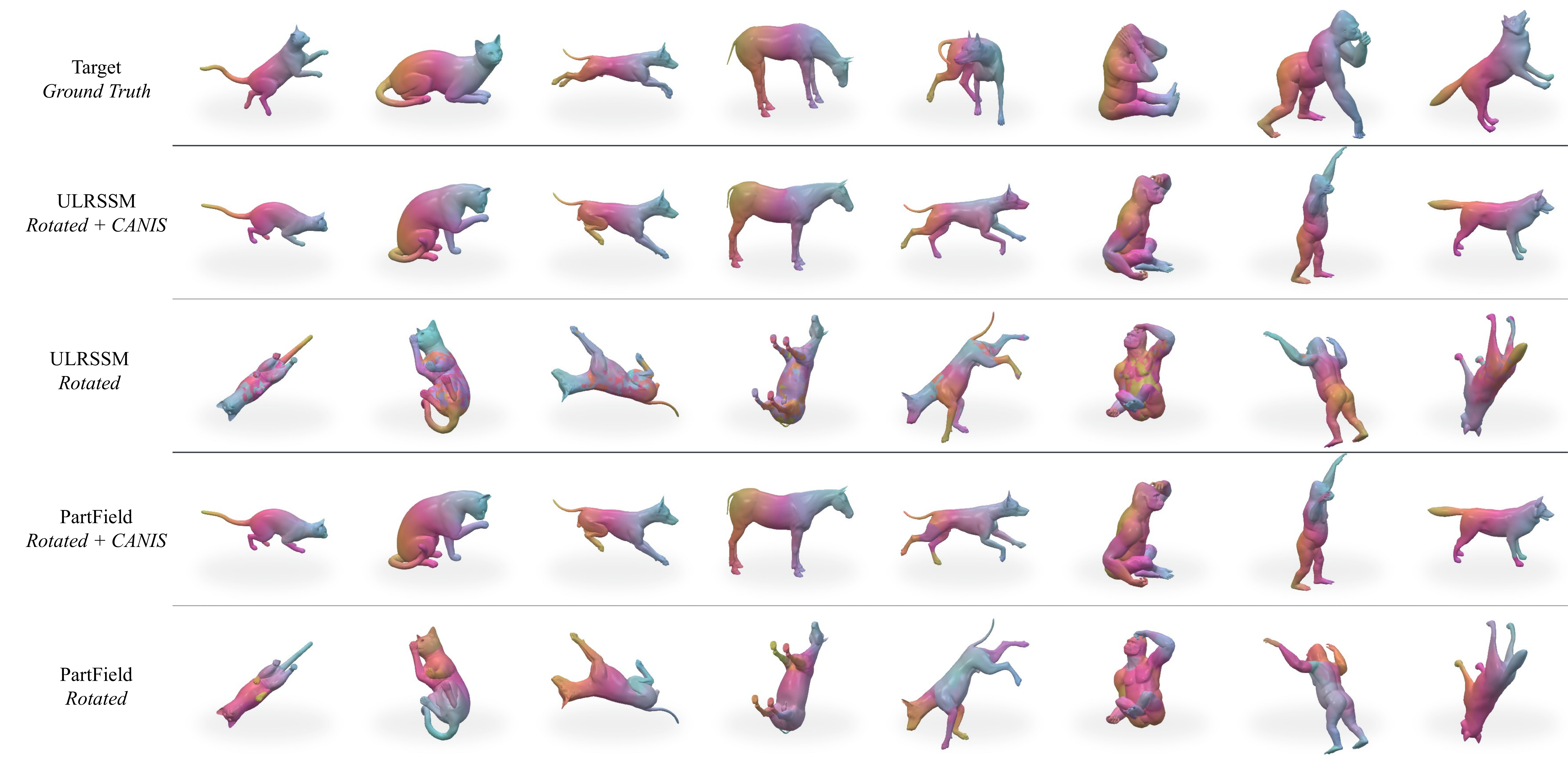}
    \caption{
    Qualitative dense correspondence results on TOSCA. The first row shows the target shapes with ground truth correspondence colors. For method ULRSSM~\cite{ulrssm} and PartField~\cite{partfield}, the result after CANIS canonicalization is shown above the result obtained directly from the same rotated input. Consistent colors on corresponding anatomical regions indicate accurate matching.}
   \label{fig:dense_matching}
\end{figure*}

\vspace{0.5em}
\mypara{3D Dense Correspondence} 
We evaluate whether CANIS improves dense correspondence on 286 ordered within-class pairs from the 41-shape TOSCA animal subset. Each rotated pair is matched using ULRSSM~\cite{ulrssm} and PartField~\cite{partfield} before and after CANIS canonicalization. Geodesic metrics use fixed deterministic samples of 256 points for ULRSSM and 512 points for PartField, while exact accuracy is computed over all vertices. We therefore compare the two input conditions within each matcher. As reported in Table~\ref{tab:dense_matching}, CANIS increases PCK@0.10 from 59.69\% to 99.01\% for ULRSSM and from 11.99\% to 67.46\% for PartField, while reducing their geodesic errors from 23.62 to 1.83 and from 57.72 to 16.57, respectively. The remaining metrics show consistent improvements. Fig.~\ref{fig:dense_matching} further shows that PartField frequently reverses head and tail correspondences under relative rotations approaching \(180^\circ\). CANIS reduces head-tail reversals and improves all reported correspondence metrics, although exact vertex accuracy remains low for both matchers.

\begin{figure*}[!t]
    \centering
    \includegraphics[width=\linewidth]
        {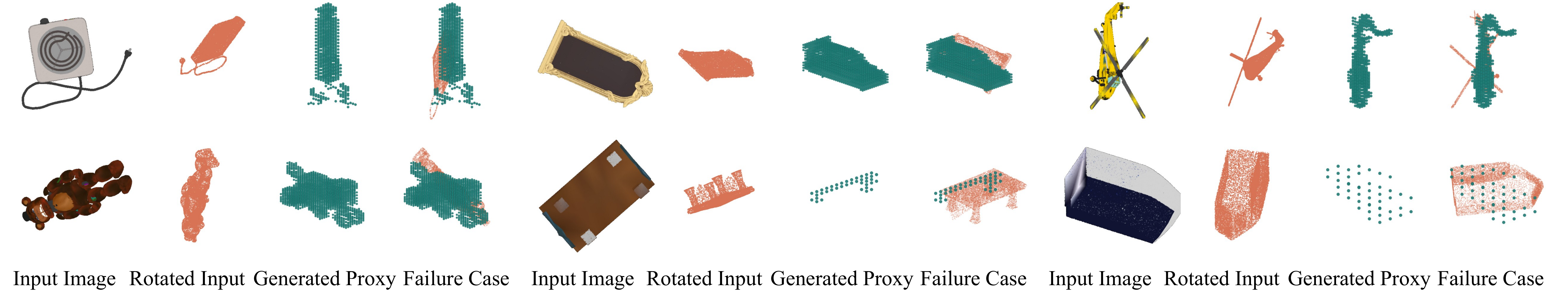}
    \caption{Representative CANIS failures. Each group shows the selected input image, the rotated input, the generated proxy, and the final canonicalization result. These examples contain insufficient orientation cues, proxy mismatch, or ambiguous correspondences, which can lead to an incorrect semantic orientation.}
    \label{fig:failure_cases}
    \vspace{-0.3cm}
\end{figure*}

\begin{table*}[t]
  \centering
  \caption{
  Dense correspondence performance on 286 ordered TOSCA animal pairs. The same SO(3) rotated inputs are evaluated directly (Rotated) and after CANIS canonicalization (Canonical). Results are averaged over all pairs. AUC, PCK, and exact accuracy are reported as percentages, while normalized geodesic error is multiplied by 100. Higher values are better except for geodesic error.
  }
  \label{tab:dense_matching}

  \footnotesize
  \setlength{\tabcolsep}{4.8pt}
  \renewcommand{\arraystretch}{1.12}

  \begin{tabular}{@{}llcccccc@{}}
    \toprule
    Method & Input
    & AUC@$0.10$ \(\uparrow\)
    & PCK@$0.01$ \(\uparrow\)
    & PCK@$0.05$ \(\uparrow\)
    & PCK@$0.10$ \(\uparrow\)
    & \shortstack{Geo. Err.\\(\(\times 100\))} \(\downarrow\)
    & \shortstack{Exact\\Acc.} \(\uparrow\) \\
    \midrule

    \multirow{2}{*}{ULRSSM~\cite{ulrssm}}
    & \cellcolor{oursrow}Canonical
    & \cellcolor{oursrow}$\textbf{83.95}$
    & \cellcolor{oursrow}$\textbf{50.37}$
    & \cellcolor{oursrow}$\textbf{93.74}$
    & \cellcolor{oursrow}$\textbf{99.01}$
    & \cellcolor{oursrow}$\textbf{1.83}$
    & \cellcolor{oursrow}$\textbf{14.02}$ \\

    & Rotated
    & $47.03$ & $24.05$ & $52.59$ & $59.69$ & $23.62$ & $3.98$ \\

    \addlinespace[2pt]
    \midrule

    \multirow{2}{*}{PartField~\cite{partfield}}
    & \cellcolor{oursrow}Canonical
    & \cellcolor{oursrow}$\textbf{39.71}$
    & \cellcolor{oursrow}$\textbf{6.30}$
    & \cellcolor{oursrow}$\textbf{43.82}$
    & \cellcolor{oursrow}$\textbf{67.46}$
    & \cellcolor{oursrow}$\textbf{16.57}$
    & \cellcolor{oursrow}$\textbf{1.85}$ \\

    & Rotated
    & $5.51$ & $0.39$ & $5.32$ & $11.99$ & $57.72$ & $0.07$ \\

    \bottomrule
  \end{tabular}
  
\end{table*}

\subsection{Failure Analysis}
Fig.~\ref{fig:failure_cases} presents representative failures of CANIS. These cases mainly arise from insufficient orientation-discriminative cues, substantial proxy--input discrepancies, or ambiguous semantic anchors caused by severe incompleteness and approximate symmetry. The resulting correspondences may support a geometrically plausible yet semantically incorrect canonical orientation. These failure modes motivate the broader discussion in Sec.~\ref{sec:limitations}.

\subsection{Efficiency Analysis}
We separately profile CANIS on a single NVIDIA RTX A6000 GPU. With 518$\times$518 renderings, a median input size of 163,840 points, and 8 sparse-structure and 8 SLAT sampling steps, the full pipeline takes 22 s per object and uses 11.56 GB of peak GPU memory. TRELLIS-OA sampling and PCAR take 15 s and 6 s, respectively. The remaining second is attributed to preprocessing and other overhead. 

\subsection{Limitation and Discussion\label{sec:limitations}}
CANIS uses informative view selection and shape guidance during proxy synthesis to reduce discrepancies between proxy and input geometry. It further filters the proxy support according to silhouette consistency and applies semantic anchor gating and geometric consensus to suppress ambiguous correspondences. These designs improve robustness in the evaluated settings, although two failure modes remain under extreme conditions. Severe occlusion or foreshortening can leave insufficient evidence for faithful proxy synthesis, weakening subsequent registration. Symmetric or repetitive regions with weak appearance cues may also produce similar attention responses and ambiguous anchor assignments, thereby reducing orientation stability.

\section{Conclusion}

We introduced CANIS, which canonicalizes an input object by registering it to an instance-specific proxy generated in a canonical orientation. Shape guidance reduces proxy-input geometry differences, while image-induced semantic anchors restrict 3D correspondence search. CANIS improves rotation and category consistency over the evaluated baselines without additional canonicalization-specific training. The same preprocessing also recovers performance lost to arbitrary rotations in the evaluated classification, part-segmentation, and dense-correspondence settings. Qualitative results on partial observations and OmniObject3D show the current transfer behavior. CANIS improves consistency across input rotations and category instances over existing methods, without fine tuning, or adaptation specific to the evaluation benchmarks.

\bibliographystyle{IEEEtran}
\IfFileExists{ieee/references.bib}{%
  \bibliography{ieee/references}%
}{%
  \bibliography{references}%
}

\begin{IEEEbiography}[{\includegraphics[width=1in]{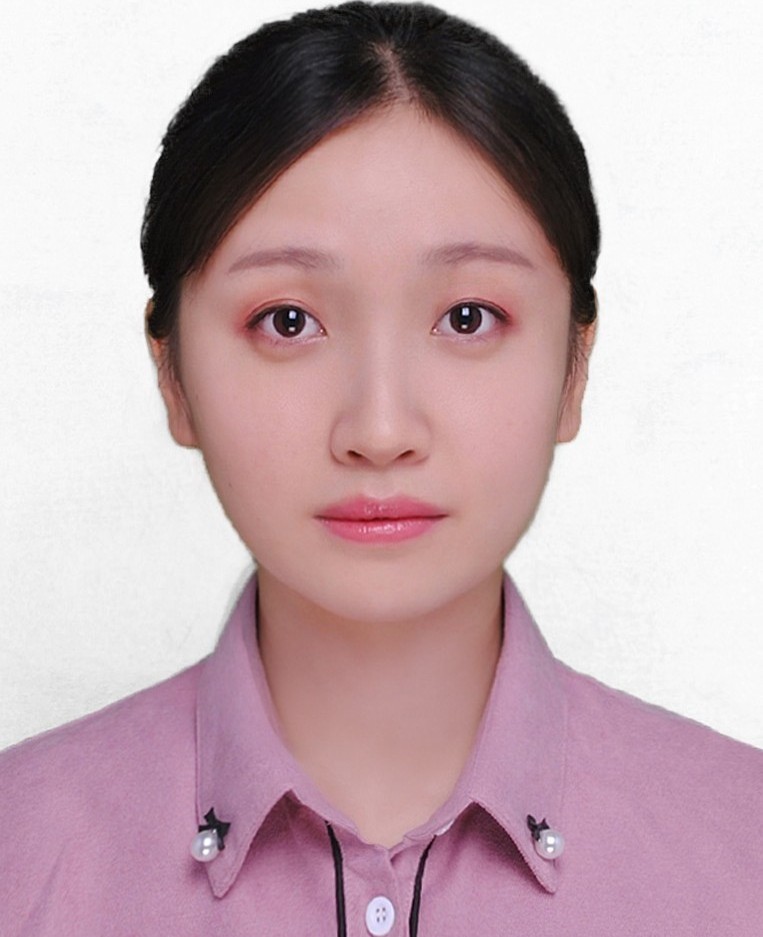}}]{Kendong Liu} is currently a PhD student with the Department of Computer Science, City University of Hong Kong. She received the BE degree from Xidian University, Xi’an, China, in 2020, and the MS degree from Xidian University, Xi’an, China, in 2023. Her research interests include generative models and 3D vision.
\end{IEEEbiography}

\begin{IEEEbiography}[{\includegraphics[width=1in]{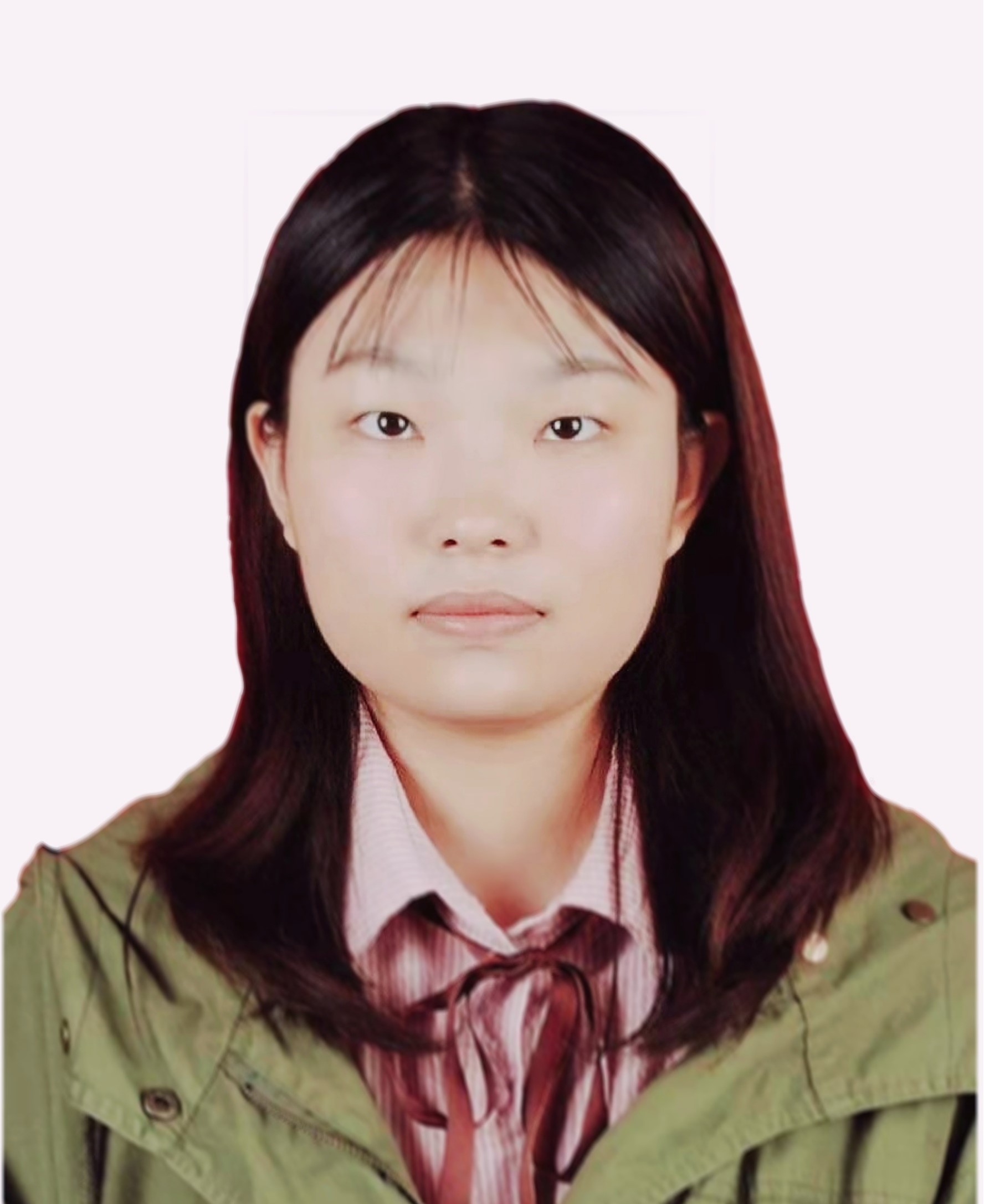}}]{Yuxin Yao} is currently a post-doctoral researcher with the Department of Computer Science, City University of Hong Kong. She received the BE degree from the University of Electronic Science and Technology of China, Chengdu,
China, in 2018, and the PhD degree from the University of Science and Technology of China, Hefei, China, in 2023. Her research interests include 3D vision and computer graphics.
\end{IEEEbiography}

\begin{IEEEbiography}[{\includegraphics[width=0.9in]{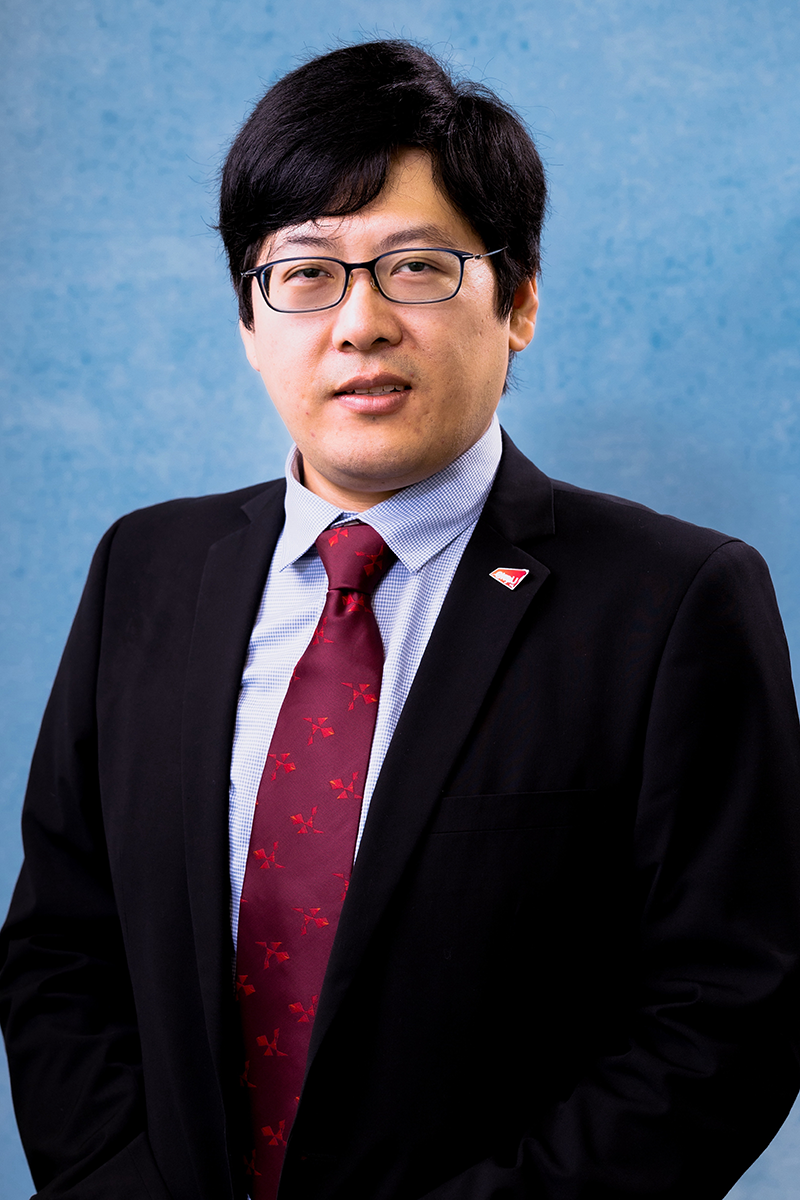}}]{Junhui Hou} is a Professor with the Department of Computer Science, City University of Hong Kong. His research interests include multidimensional visual computing, such as light field, hyperspectral, geometry, and event data. He received the Early Career Award from the Hong Kong Research Grants Council in 2018,  the Excellent Young Scientists Fund from NSFC in 2024, and the IEEE TIP Best Paper Award in 2025. He is serving as a Senior Area Editor for IEEE TIP and an Associate Editor for IEEE TVCG and TMM. He served as an Associate Editor for IEEE TIP and TCSVT.
\end{IEEEbiography}

\end{document}